\documentclass[twoside]{article}

\usepackage[accepted]{aistats2024}

\usepackage[round]{natbib}

\usepackage{tikz}
\usepackage{pgfplots}
\pgfplotsset{compat=newest}
\usetikzlibrary{fit}
\usepackage[utf8]{inputenc} %
\usepackage[T1]{fontenc}    %
\usepackage{hyperref}       %
\usepackage{url}            %
\usepackage{booktabs}       %
\usepackage{amsfonts}       %
\usepackage{nicefrac}       %
\usepackage{microtype}      %
\usepackage{xcolor}
\definecolor{MCcolor1}{RGB}{253, 231, 37}
\definecolor{MCcolor2}{RGB}{53, 183, 121}
\definecolor{MCcolor3}{RGB}{49, 104, 142}
\definecolor{MCcolor4}{RGB}{68, 1, 84}
\definecolor{alphacolor1}{RGB}{237, 121, 83}
\definecolor{alphacolor2}{RGB}{156, 23, 158}
\definecolor{alphacolor3}{RGB}{13, 8, 135}
\usepackage{amsfonts, amsmath, amssymb, amsthm, dsfont} %
\usepackage{todonotes}
\usepackage[noend]{algpseudocode}
\usepackage{algorithm}
\usepackage{comment}

\theoremstyle{plain}
\newtheorem{theorem}{Theorem}

\newtheorem{corollary}[theorem]{Corollary}
\theoremstyle{remark}
\newtheorem{remark}[theorem]{Remark}

\newcommand{\R}{\mathbb{R}}
\newcommand{\GP}{\mathcal{GP}}

\begin{document}

\runningauthor{Tebbe, Zimmer, Steland, Lange-Hegermann, Mies}

\twocolumn[

\aistatstitle{Efficiently Computable Safety Bounds for Gaussian Processes in Active Learning}

\aistatsauthor{Jörn Tebbe\And Christoph Zimmer}

\aistatsaddress{OWL University of Applied Sciences and Arts \And Bosch Center for Artificial Intelligence}

\aistatsauthor{Ansgar Steland \And Markus Lange-Hegermann}

\aistatsaddress{RWTH Aachen University \And OWL University of Applied Sciences and Arts}

\aistatsauthor{Fabian Mies}
\aistatsaddress{TU Delft}
]

\begin{abstract}
  Active learning of physical systems must commonly respect practical safety constraints, which restricts the exploration of the design space. 
    Gaussian Processes (GPs) and their calibrated uncertainty estimations are widely used for this purpose.
    In many technical applications the design space is explored via continuous trajectories, along which the safety needs to be assessed. 
    This is particularly challenging for strict safety requirements in GP methods, as it employs computationally expensive Monte-Carlo sampling of high quantiles.
    We address these challenges by providing provable safety bounds based on the adaptively sampled median of the supremum of the posterior GP. 
    Our method significantly reduces the number of samples required for estimating high safety probabilities, resulting in faster evaluation without sacrificing accuracy and exploration speed.
    The effectiveness of our safe active learning approach is demonstrated through extensive simulations and validated using a real-world engine example.
\end{abstract}

\begin{figure*}[t]
\input{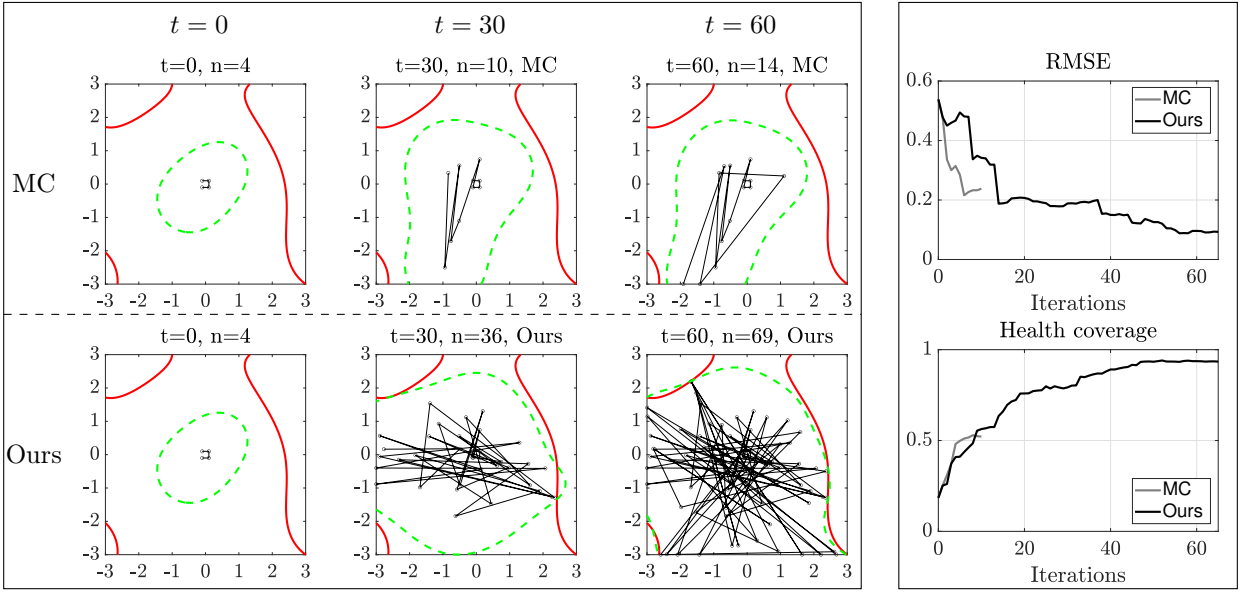}
    \caption{By providing better estimates, we obtain accurate error bounds with much fewer MC samples.
    This reduction in computation time for the safety evaluation allows more time to obtain measurements for Safe Active Learning.
    The three left columns present visual representations of a Safe Active Learning task. Each column corresponds to a different algorithm runtime $t$, along with the respective number of training points $n$, resulting in $n-1$ explored trajectories. In these plots, the green dashed region marks the area classified as safe by the GP, while the space outside the red boundary indicates the ground truth unsafe region.
    The rightmost column provides comparisons of two crucial metrics, contingent on the number of iterations. These comparisons underscore the superiority of our approach in enhancing the effectiveness of the Safe Active Learning process by allowing more iteration in the same time.
    }
    \label{fig:contribution}
\end{figure*}

\section{INTRODUCTION}

Active learning is a machine learning technique that involves selecting the most informative examples from a large unlabeled dataset and requesting their labels from an oracle, e.g., a human annotator or a costly experiment, to improve the performance of a learning algorithm \citep{Settles09,tharwat2023survey}.
The goal of active learning is to reduce the number of labeled examples needed to achieve high accuracy. %

In many engineering scenarios, the next experiment must not only be informative, but also adhere to \emph{practical safety constraints} \citep{sui2018stagewise,Berkenkamp16,Baumann21}. 
For example, we consider the control of a high-pressure fluid system for fuel injection in combustion engines, see Section \ref{subsection_railpressure}. 
The experimental conditions need to be chosen such that a critical pressure threshold is not exceeded. 
Challengingly, the exact effect of the controls on the pressure in the system is unknown and needs to be learned simultaneously.
Safe active learning aims to balancing the trade-off between exploration performance and safety, ensuring that the selected examples are not only informative but can also be obtained securely.

An established approach in safe active learning is to model the unknown functional relation via Gaussian processes (GPs) \citep{rasmussen2006gaussian}.
This allows for Bayesian uncertainty quantification and, hence, for informed decisions about where to sample next \citep{schreiter2015safe,zimmer2018safe,Li2022}.
This makes GPs a powerful tool for maximizing physical experiments' information while minimizing risk.

A typical additional challenge is exploring \emph{dynamical systems}, where exploration along trajectories instead of single datapoints is necessary.
In our engine example a controller continuously adapts engine speed and rail pressure, and similarly the path in robot exploration needs to be safe.
In such applications, the entire trajectory needs to be safe.

GPs are again attractive for exploration along trajectories.
They naturally induce a univariate GP as posterior of the safety constraint along the trajectory.
Now, the safety probability is the probability of this univariate GP being safe which is analytically intractable.
The state of the art approach is to \emph{estimate the safety of paths} by generating enough Monte-Carlo (MC) samples from the GP on a finite set of points on the path \citep{zimmer2018safe,zimmer2020}.

However, this has a major drawback:
obtaining high safety guarantees requires a large number of MC samples, as one needs enough samples in the tails of the distribution. 
This exposes a trade-off between the safety of the trajectory, the confidence in the safety assessment, and the computational costs.
The latter is especially important if decisions need to be taken quickly, as otherwise costly measurement equipment and staff is idly waiting \citep{sandmeier2022optimization,thewes2016efficient}.

This paper tackles this trade-off by a novel, comparatively tight, and computationally efficient algorithms to compute provable \emph{probabilistic upper bounds} on GP maxima evaluated on a discretization.
We develop a variant of the Borell-TIS bound, make it well-suited for practical applications, and adapt this bound to allow for non-centered GPs.
The Borell-TIS inequality allows us to draw conclusions about the far tails of the posterior distribution based on the median and the maximal variance, which can be estimated adaptively and reliably with comparatively few MC samples.
Our algorithm uses this inequality adaptively, and hence achieves high precision with minimal computational overhead, enabling faster and more efficient learning.
This paper makes the following contributions:
\begin{itemize}
    \item We reduce the \emph{safety assessment for a non-centered GP} to a centered GP, see Remark~\ref{remark_transform_gp} and Figure~\ref{figure_centered_GP}.
    \item We propose a computationally efficient method for the safety assessment using an \emph{adaptive MC sampling scheme}.
    \item We rigorously prove in Section~\ref{section_theory} and empirically demonstrate on various examples in Section~\ref{section_examples} the \emph{safety guarantees} of our approach.
    \item We demonstrate that our approach is significantly faster to compute than state-of-the-art safe active learning techniques, hence it enables \emph{more exploration in the same time} (see Section \ref{section_examples}).
\end{itemize}

\section{PRELIMINARIES}

\subsection{Gaussian processes (GPs)}\label{sec:recap}

A Gaussian process (GP) $g=\GP(\mu,k)$ is a stochastic process characterized by its mean function $\mu \colon \mathbb{R}^d \to \mathbb{R}$ and covariance function $k \colon \mathbb{R}^d \times \mathbb{R}^d \to \mathbb{R}$.
Conditioning $g=\GP(\mu,k)$ on a dataset $(x,y)\in\R^{n\times(d+1)}$ yields a posterior GP with mean and covariance functions
\begin{align*}
    \mu^*(x^*) &= \mu(x^*)+k(x^*, x) (K + \sigma_n^2 I)^{-1} y \\
    \Sigma(x_1,x_2) &= k(x_1, x_2) - k(x_1, x) (K + \sigma_n^2 I)^{-1} k(x, x_2) 
\end{align*}
with covariance matrix $K=(k(x_i,x_j))_{i,j} \in \mathbb{R}^{n\times n}$, and $k(x^*, x)\in\R^{1\times n}$, $k(x_1, x)\in\R^{1\times n}$, $k(x, x_2), k(x, x^*)\in\R^{n\times 1}$ and noise variance $\sigma_n^2$ \citep{rasmussen2006gaussian}.
In practice, GPs are parameterized by hyperparameters $\theta \in \mathbb{R}^p$.
These hyperparameters are adapted to data by minimizing the negative log-likelihood $p(y|x,\theta)$. %
We use the squared exponential covariance function 
\[k_{\text{SE}}(x_1,x_2)=\sigma_f^2\exp\left(-\frac{1}{2}\frac{(x_1-x_2)^2}{\ell^2}\right)\]
for the GP priors.

Sampling a GP $g=\GP(\mu,k)$ at a finite number of points $x_1^*,\ldots,x_m^*\in\R^d$ amounts to sampling from the $m$-dimensional Gaussian distribution $\mathcal{N}(v,K^*)$ with $v_i=\mu(x_i^*)$ and $(K^*)_{i,j}=k(x_i^*,x_j^*)$ for $i,j\in\{1,\ldots,m\}$.
The computational costs for generating $M$ samples consist of $\mathcal{O}(m^3) $ operations for a preliminary Cholesky decomposition of the covariance matrix $K^* $ and $\mathcal{O}(Mm^2) $ operations to simulate the samples, in addition to the cost of computing the posterior.

\subsection{Safe Active Learning}

Safe Active Learning selects sample locations $x_1, \ldots, x_n$ that maximize information content, constrained on safety \citep{schreiter2015safe,zimmer2018safe,Li2022}. Entropy is a measure of information that is particularly suited for GPs as the entropy of a new point $x^*$ is in monotonous bijection to its predictive variance $\sigma^2(x^*)$. Therefore, the core of safe active learning is a constrained optimization problem:
\begin{gather*}
    x_{n+1} = \text{argmax}_{x^*\in\mathcal{X}}\ \sigma(x^*)
    \quad\text{s.t.}\quad P_{\text{unsafe}}(x^*) \leq \alpha    
\end{gather*}
where a small $0 < \alpha \le 1$ denotes the maximal desired probability of unsafety and $\mathcal{X}\subseteq \mathbb{R}^d$ is the operation area of the system.
We consider the usual case that the safety of a point $x^*$ characterized in terms of a (unknown) safety indicator function $f:\R^d:\to\R$.
That is, an operational setting $x^*$ is safe if $z=f(x^*) \geq z_{\min}$.
By shifting the mean value accordingly, we may set $z_{\min} =0$ without loss of generality. 
We assume the safety indicator to be experimentally measurable, so that we can model it via a posterior GP $\widehat{f} \sim GP(\widehat{\mu},\widehat{\Sigma})$, where the posterior parameters $\widehat{\mu}$ and $\widehat{\Sigma}$ depend on the previously explored samples $x_1,\ldots,x_n$ and their evaluations $z_i=f(x_i)$, as described in Section \ref{sec:recap}.
Then we denote by $P_{\text{unsafe}}(x^*)$ the posterior probability
\begin{align}
    P_{\text{unsafe}}(x^*) = 1 - \int_{z \geq 0} \mathcal{N}(z; \widehat{\mu}(x^*),\widehat{\Sigma}(x^*,x^*)) dz.
    \label{P_unsafe}
\end{align}

In case of active learning in \emph{dynamic systems} \citep{zimmer2018safe}, the exploration is usually conducted along parameterized trajectories $\tau(t)_{t\in[0,1]}$ instead of points $x^*$, e.g.\ the trajectory $\tau$ can be a linear ramp leading to some end point of interest.
The active learning task then consists in choosing a sequence $\tau_1, \tau_2,\ldots$ of trajectories which maximize information content, constrained by the safety requirement $P_{\text{unsafe}}\leq \alpha$.
A measurement is conducted at the endpoint of the trajectory $\tau$, and information is then measured with the posterior GPs $\widehat{f}$ predictive variance $\widehat{\sigma}( \tau(1))$.
The considered probability of the trajectory being unsafe is 
\begin{align}
    P_{\text{unsafe}}(\tau)  = P\left( \inf_{t\in [0,1]} Z_t \leq 0 \right),
    \label{P_unsafe_GP}
\end{align}
where $Z_t$ is a sample of the posterior GP. 

In the same framework, it is also possible to conduct measurements along the trajectory at locations $\tau(t_1), \ldots, \tau(t_m)$ for $t_1,\ldots,t_m\in [0,1]$. 
The information of these measurements may be expressed in terms of the predictive covariance matrix $\widehat{\Sigma}( \tau(t_1), \ldots, \tau(t_m))$, and quantified via its trace or determinant. 

\section{RELATED WORK}

Estimating bounds for a GP is a crucial task for exploration in safety critical environments. Other approaches in the literature consider a bound of the RKHS norm of the safety function in order to create confidence intervals. While \cite{sui2018stagewise} propose to additionally use estimated Lipschitz-constants of the safety function, \cite{bottero2022information} overcomes this practically strong assumption.
In \cite{lederer2019}, the authors provide uniform error bounds based on Lipschitz constants estimations.
\cite{schreiter2015safe} proposes a method to obtain pointwise safety in safe active learning. 
All these works consider pointwise safety instead of safety on continuous trajectories, \cite{zimmer2018safe} extend safety consideration to trajectories.
This is improved in \cite{zimmer2020} by an adaptive discretization scheme.
Our novel approaches are compatible with both these papers and for simplicity we compare ourselves to the former one.

Moreover, \cite{cardelli2019robustness} consider safety for compact sets to detect adversarial attacks on the data. For this purpose they use a variant of the Borell-TIS inequality \citep{Adler2007} which bounds the mean of the supremum of a GP using Dudley's theorem \citep{dudley1967sizes}. We show in Section \ref{subsection_toy_example}, that these bounds are inferior, compared to our proposed method.

While our experiments use standard GPs with squared exponential covariance function, our methods directly extend to usual variants and approximations to GPs.
This includes any separable covariance function, including specific ones constructed from kernel search \citep{duvenaud2013structure,Bitzer2022}, 
geometry \citep{borovitskiy2020matern},  differential equations \citep{besginow2022constraining,harkonen2022gaussian}, %
for high dimensional modeling  \citep{duvenaud2011additive},
kernels building on Fourier frequencies \citep{lazaro20sparse},
or symmetry \citep{holderrieth2021equivariant}. 
Furthermore, this includes GPs for big data, be it via variational approximations \citep{titsias2009variational,hensman2013gaussian,hensman2017variational}, via kernel approximations \citep{wilson2015kernel}, or improved linear algebra \citep{gardner2018gpytorch,wang2019exact}.

The safety assessment in \eqref{P_unsafe} reduces to a Gaussian integral under linear constraints which has been considered by \cite{genz1992numerical}. This method has been extended to high dimensional integrals ($m > 100$) of mainly small areas \citep{botev2017normal,gessner2020integrals}. These methods use variants of Monte-Carlo sampling, but prior experiments indicate that they have inferior performance as our proposed methods.

Adaptive stopping Monte-Carlo schemes have been used in the literature for estimating statistical quantities \citep{mnih2008empirical}. Our stopping scheme extends via exploiting an additional binomial structure.

\section{UNIFORM TAIL BOUNDS FOR GAUSSIAN PROCESSES}\label{section_theory}

When exploring the experimental space along a trajectory $(\tau(t))_{t\in[0,1]}\in\R^d$, we want to be reasonably certain that a safety indicator $f(\tau(t))$ does not fall below a critical threshold $z_{\min} = 0$.  
Denote the trajectories of the posterior distribution for $f(\tau(t))_{t\in[0,1]}$ by $(Z_t)_{t\in[0,1]}$, i.e.\ 
$Z_t$ is a GP with one-dimensional input, mean function $\mathbb{E}(Z_t)=\mu_t\in\R, t\in[0,1]$, and covariance function $\text{Cov}(Z_s, Z_t) = \Sigma_{s,t}$, derived according to the formulas in Section \ref{sec:recap}.
Hence, the (posterior) probability $P_{\text{unsafe}}$ of the trajectory $\tau$ being unsafe is given by \eqref{P_unsafe_GP}.
For computational purposes, the continuous trajectory $\tau(t)$ needs to be discretized by considering only a subset $T\subset [0,1]$, such that
\begin{align*}
    P_\text{unsafe}(\tau) %
    &\approx P\left( \inf_{t\in T} Z_t \leq 0 \right) =: P^*(\tau). 
\end{align*}
Typically, $T=\{t_1,\ldots, t_m\}$ is a finite set for computational purposes.
However, we want to emphasize that our adaptive sampling scheme and the analytical bounds presented in Section \ref{subsec-analytical} also hold for the continuous case $T=[0,1]$.

For safe exploration, we want to accept the trajectory $\tau(t)_{t\in T}$ as being safe only if $P^*(\tau) \leq \alpha$, for some small $\alpha\in (0,1]$.
This section describes reliable upper bounds on $P^*$ which should be (i) fast to evaluate computationally and (ii) as sharp as possible.

\subsection{Adaptive Monte-Carlo Sampling (AMC)}\label{subsection_adaptive_MC}

The state of the art %
chooses a potentially non-equidistant discretization of $[0,1]$ %
given by $0\le t_1<\ldots<t_m\le 1$ and simulate a potentially large number, $M$, of trajectories $Z_{t,1},\ldots, Z_{t,M}$ evaluated only on the discretization $T=\{ t_1,\ldots, t_m \}$.
The computational cost is dominated by $\mathcal{O}(Mm^2)$ for large enough $M$, see Section \ref{sec:recap}.
We estimate the probability of unsafe trajectories by their proportion in the sampled trajectories
\[ \widehat{P}_{\textsc{MC}} (\tau, M) = \frac{1}{M} \sum_{i=1}^M \mathbf{1} \left( \min_{j=1,\ldots,m} Z_{t_j, i} \leq 0\right). \]

How many MC samples $M$ are necessary? 
Since $\text{Var}(\widehat{P}_{\textsc{MC}}) \leq P^*/M$, the relative error compared to the safety threshold is of the order 
\[|\widehat{P}_{\textsc{MC}} - P_{\textsc{MC}}|/\alpha = \mathcal{O}(\sqrt{P^*}/\sqrt{M\, \alpha}).\]
Thus, for trajectories which are barely safe, $P^*\approx \alpha$, we should perform $M\asymp \alpha^{-1}$ MC iterations. 
That is, for strict safety requirements $\alpha\approx 0$, the MC approach is computationally expensive.
Determining the exact number of required MC samples is non-trivial: 
If $P^*\ll \alpha$, i.e.\ if the trajectory is very safe, then few samples suffice as the variance of $\widehat{P}_{\textsc{MC}}$ is small. 
If, on the other hand, $P^*\gg \alpha$, it is also sufficient to draw rather few samples, as the mean of $\widehat{P}_{\textsc{MC}}$ is far away from the decision boundary. 
Specifically in the critical regime $P^*= \alpha + O(\delta)$ for some small $|\delta|$ it is hard to decide whether the trajectory $\tau$ is indeed safe or unsafe; the smaller $|\delta|$, the more samples are needed. 
Unfortunately, given a candidate trajectory $\tau$, we do not know $P^*$ in advance. 

We suggest to determine the sample size adaptively: generate the MC samples sequentially and perform an online test for the hypothesis $H_0: P^*(\tau) \geq \alpha$, and to stop sampling as soon as $H_0$ is rejected, classifying the trajectory as safe. 
At the same time, we stop when $H_0':P^*(\tau)\leq \alpha$ is rejected, classifying the trajectory as unsafe.
We suggest to sequentially increase the sample size $M_1 < M_2 < \ldots$, and to stop sampling at step $r^*$ with sample size $M_{r^*}$ for
\begin{align*}
	r^* = \inf \Big\{r\;:\; &\widehat{P}_{\textsc{MC}}^+(\tau, M_r, r, \epsilon, \alpha) < \alpha \;\text{or}\; \\ &\widehat{P}_{\textsc{MC}}^-(\tau, M_r, r, \epsilon, \alpha) > \alpha)\Big\}, \\
    \widehat{P}_{\textsc{MC}}^+(\tau, M_r, r, \epsilon, \alpha):=&\widehat{P}_{\textsc{MC}}(\tau, M_r) + \sqrt{\alpha (1-\alpha)} c_r, \\
    \widehat{P}_{\textsc{MC}}^-(\tau, M_r, r, \epsilon, \alpha):=&\widehat{P}_{\textsc{MC}}(\tau, M_r) - \frac{c_r^2}{4} -  c_r \sqrt{\alpha}, \\
 \text{with }  c_r &= \sqrt{\frac{2}{M_r} \left|\log \frac{6\epsilon}{\pi^2 r^2}\right|}
\end{align*}
where $\epsilon>0$ should be small.
Stopping due to $\widehat{P}_{\textsc{MC}}^+$ resp.\ $\widehat{P}_{\textsc{MC}}^-$ classifies $\tau$ as safe resp.\ unsafe.
Indeed, this procedure can control the probability of falsely classifying an unsafe trajectory as safe and vice versa.

\begin{theorem}\label{thm:adaptive}
    Let $Q$ denote the probability w.r.t.\ the MC sampling and let $\epsilon\in(0,1)$.
    If ${P}^*(\tau)\geq \alpha$ then
    \begin{align*}
        Q\left( \exists r\in\mathbb{N}: \, \widehat{P}_{\textsc{MC}}^+(\tau, M_r, r, \epsilon,\alpha) < \alpha \right) \leq \epsilon
    \end{align*}
    and if ${P}^*(\tau)\leq \alpha$, then
    \begin{align*}
        Q\left( \exists r\in\mathbb{N}: \, \widehat{P}_{\textsc{MC}}^-(\tau, M_r, r, \epsilon, \alpha) > \alpha \right) \leq \epsilon.
    \end{align*}
\end{theorem}

We call this method Adaptive Monte-Carlo (AMC).
Since $c_r\to 0$ and $\widehat{P}_{\textsc{MC}}(\tau, M_r)\to P^*$, the stopping time $r^*$ is almost surely finite if $P^*\neq \alpha$. 
Theorem~\ref{thm:adaptive} guarantees the above method decides correctly with probability $1-\epsilon$ if $P^*\neq \alpha$; we may choose $\epsilon$ small.
On the other hand, Theorem \ref{thm:adaptive} also implies that $P(r^*=\infty)\geq 1-\epsilon$ for the edge case $P^*=\alpha$. 
Thus, in practice, we impose an upper bound on $r^*$ and classify a trajectory as unsafe if this bound is exceeded.

\subsection{Analytical bound}\label{subsec-analytical}

We suggest to bound $P^*$ via the Borell-TIS inequality for GPs.
It asserts the remarkable result that the supremum of a GP shifted by the mean or median of suprema of samples has subgaussian tails.

\begin{theorem}[Borell-TIS inequality]\label{thm:Borell-TIS}
    Let $X_t, t \in T$, be a centered separable GP with index set $T$, and maximal pointwise variance $\sigma^2 = \sup_{t\in T} \text{Var}(X_t)$. 
    Denote $m(X) = \mathrm{median}\left(\sup_{t\in T} X_t\right)$, $\mu(X)=\mathbb{E}\left(\sup_{t\in T} X_t\right)$ and $\Phi$ as the standard Gaussian distribution function.
    Then, for any $u\geq 0$,
    \begin{align}
        P\left[\sup_{t\in T} X_t > u + m(X)\right] &\leq [1-\Phi(u/\sigma)] \label{eqn:Borell-1} \tag{B.1} \\ 
        &\leq \tfrac{1}{2} \exp(-\tfrac{1}{2} u^2/\sigma^2), \label{eqn:Borell-2} \tag{B.2} \\
        P\left[\sup_{t\in T} X_t > u + \mu(X)\right] &\leq \exp(-\tfrac{1}{2} u^2/\sigma^2). \label{eqn:Borell-3} \tag{B.3}
    \end{align}
    
\end{theorem}

Typically, \eqref{eqn:Borell-1} is the sharpest of the bounds of Theorem~\ref{thm:Borell-TIS}; cf.\ Figure~\ref{figure_toy_comparison} for a numerical comparison.
See \cite{VanderVaart1996} for a proof of Theorem~\ref{thm:Borell-TIS}. Note that the sharpest inequality \eqref{eqn:Borell-1} is derived in the proof of Lemma A.2.2 therein.
\begin{figure*}[t]
    \begin{center}
        \includegraphics[scale=0.42]{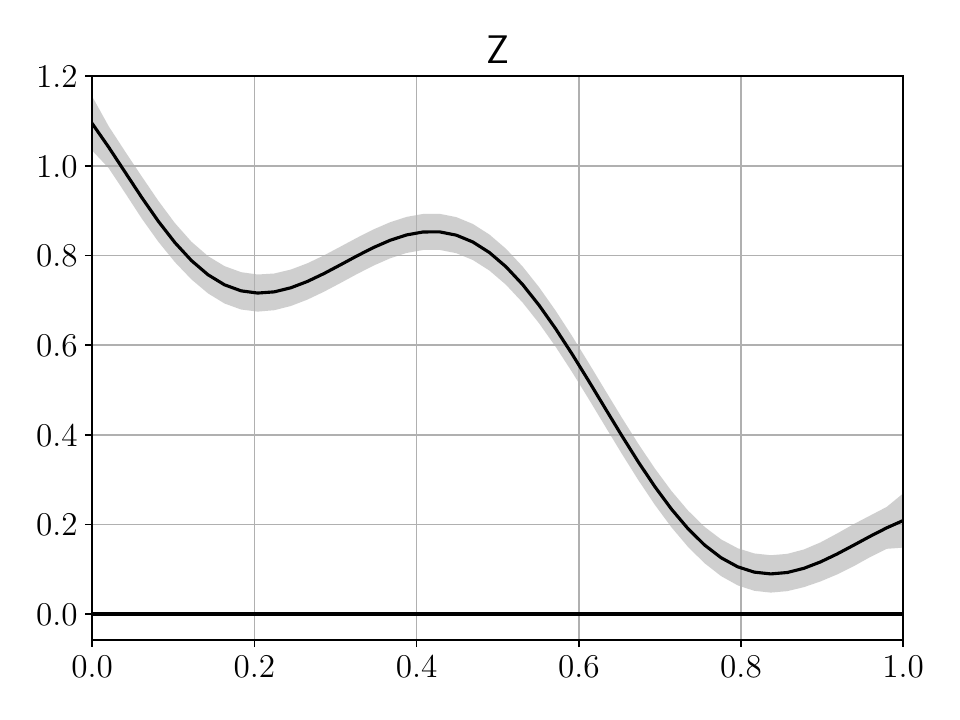}
        \includegraphics[scale=0.42]{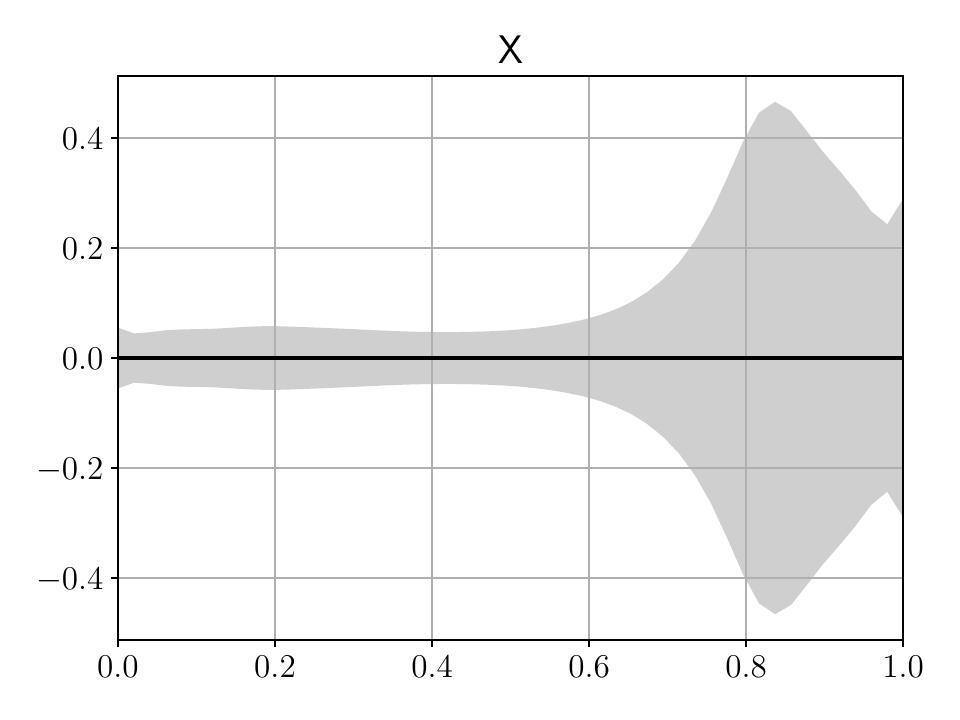}
    \end{center}
    \caption{%
        The left diagram illustrates the GP $Z_t$ from Subsection~\ref{subsection_toy_example} via its mean function and pointwise two sigma bands; the right diagram shows the corresponding centered GP $X_t$ resulting from Remark~\ref{remark_transform_gp}.
        The safety-relevant information of the mean of the original GP $Z$ moves to the covariance of the centered GP $X$.
        In particular, the variance of the centered GP $X$ rises where the mean of $Z$ approaches the safety bound zero.}
    \label{figure_centered_GP}
\end{figure*}

\begin{remark}\label{remark_transform_gp}
In our setting, Theorem \ref{thm:Borell-TIS} is not directly applicable because $Z=\GP(\mu,\Sigma)$ is usually conditioned on previously obtained data and hence of variable mean.
We remedy this technical problem as follows: suppose that $\mu_t> 0$ for all $t\in T$, since otherwise $P^*\geq \frac{1}{2}$ which is a too high risk of failure.
Then, we may rewrite
    \begin{align*}
    \inf_{t\in T} Z_t \geq 0 
    & \iff X_t:=\frac{Z_t-\mu_t}{-\mu_t} \leq 1 \quad \forall t\in T. \label{eqn:def-X} 
    \end{align*}
\end{remark}
Figure~\ref{figure_centered_GP} exemplifies this centering transformation.
Considering $X = \frac{\mu-Z}{\mu}=\GP\left(0,(s,t)\mapsto\frac{\Sigma(s,t)}{\mu(s)\mu(t)}\right)$, it suffices to derive upper tail bounds for
\[ P^* = P\left( \inf_{t\in T} Z_t \leq 0 \right) = P\left( \sup_{t\in T} X_t \geq 1 \right). \]
This discussion, together with Theorem~\ref{thm:Borell-TIS} proves the following analytical bound on $P^*$.

\begin{theorem}\label{cor:Borell-TIS}
    Let $Z_t, t \in T$, be a separable GP with index set $T$ and mean function $\mu_t > 0$. 
    Then, if $\widetilde{m} = \mathrm{median}\left(\sup_{t\in T} (\mu_t-Z_t)/\mu_t\right)\leq 1$, we have
    \begin{align*}
        P^*(\tau) &\leq P^\dagger(\tau) = 1- \Phi\left(\tfrac{1-\tilde{m}}{\tilde{\sigma}}\right)
    \end{align*}
    for $\widetilde{\sigma}^2 = \sup_{t\in T} \text{Var}(Z_t)/\mu_t^2$.
\end{theorem}

\subsection{Semi-analytical bound (AB)}

In Theorem~\ref{cor:Borell-TIS}, the median $\widetilde{m}$ and the maximal variance $\widetilde{\sigma}^2$ cannot be computed in closed form.
We approximate them via MC sampling along a discretization of $Z_t$ resp.\ $X_t$.
Thereby, we replace MC sampling of events in tails of probability distributions by much easier MC sampling of a median of the same distribution.
This is possible because the Gaussianity of the process allows us to extrapolate to the tail of the distribution using data from the center of its mass, via Theorem \ref{cor:Borell-TIS}.
As a consequence, we can drastically decrease the number of MC samples required for reliable tail bounds. %

Again, we suggest to determine the MC sample size adaptively.
To this end, we make use of exact finite sample confidence intervals for the median as follows.
Based on $M\in\mathbb{N}$ samples, denote the simulated maxima by $S_i = \max_j X_{t_j,i}$ and let $q_{\beta,M}=q_{\beta,M}(S_1,\ldots, S_M)$ the empirical $\beta$-quantile, i.e.\ the $\lfloor M\beta\rfloor$-th order statistic.
For any $M$, $r$, and $\epsilon>0$, and a confidence level $\chi=\chi(r, \epsilon)=1-\frac{6\epsilon}{\pi^2r^2}$, there exist $\beta_\pm = \beta_\pm(M,r,\epsilon)$, with $\beta_-\leq \beta_+$, such that 
\begin{align*}
    P\left( \widetilde{m}\in [q_{\beta_-,M_r}, \infty) \right) &\geq \chi, \\
    P\left( \widetilde{m}\in (-\infty, q_{\beta_+,M_r}] \right) &\geq \chi.
\end{align*}
We can use the asymptotic approximation
	$\beta_\pm \approx \frac{1}{2} \pm \Phi^{-1}(\chi) /\sqrt{4\,M}$ \citep[p.\ 144]{conover1999}.
Thus, as in Section~\ref{subsection_adaptive_MC}, we suggest to sequentially increase the sample size $M_1 < M_2 < \ldots$, and to stop sampling at step $r'$ with sample size $M_{r'}$ for 
\begin{align*}
	r' = \inf \Big\{r\;:\; 
	\widehat{P}^\dagger_{-}(M_r,r, \epsilon)> \alpha \;\text{or}\; \widehat{P}^\dagger_{+}(M_r,r,\epsilon) < \alpha \Big\}, \\
	\text{where} \qquad \widehat{P}^\dagger_{\pm}(M_r,r,\epsilon) := 1-\Phi\left(\tfrac{1-q_{\beta_\pm,M_r}}{\sigma_m}\right).
\end{align*}

\begin{theorem}\label{thm:adaptive-Borell}
	If $P^\dagger(\tau)\geq \alpha$, then for any $\epsilon\in(0,1)$
	\begin{align*}
		Q\left( \exists r\in\mathbb{N}: \, \widehat{P}^\dagger_{+}(M_r,r, \epsilon)< \alpha \right) \leq \epsilon
	\end{align*}
	and if $P^\dagger(\tau)\leq \alpha$, then
	\begin{align*}
		Q\left( \exists r\in\mathbb{N}: \, \widehat{P}^\dagger_{-}(M_r,r, \epsilon) > \alpha \right) \leq \epsilon.
	\end{align*}
	where $Q$ denotes probability w.r.t.\ the MC sampling. 
\end{theorem}
\begin{figure*}[ht]
    \begin{center}
        \includegraphics[scale=0.4]{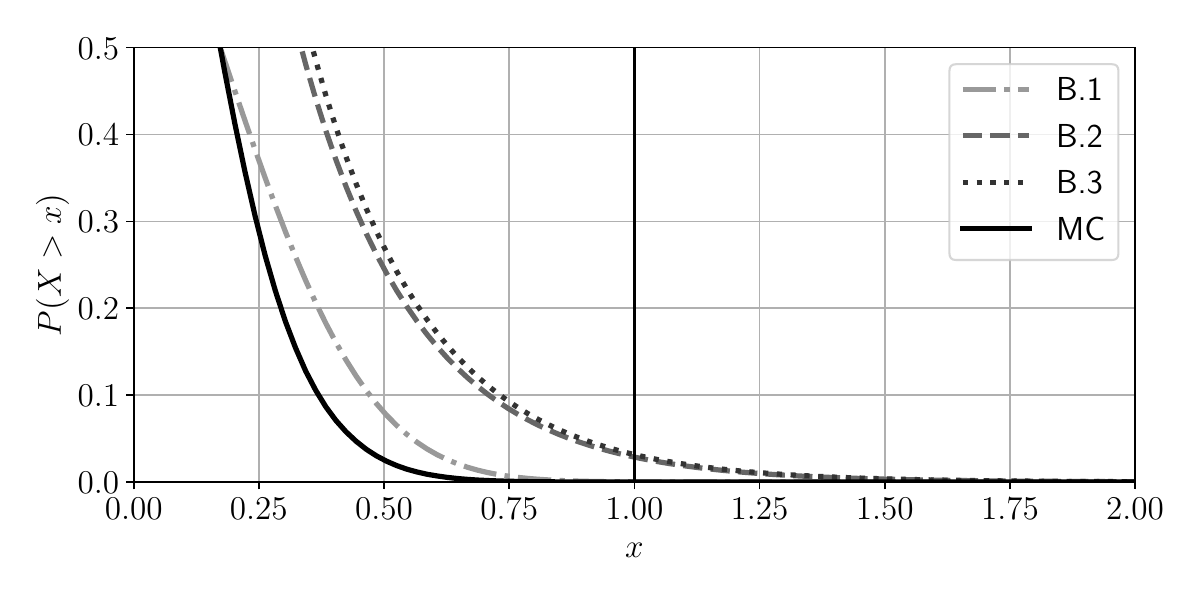}
        \includegraphics[scale=0.4]{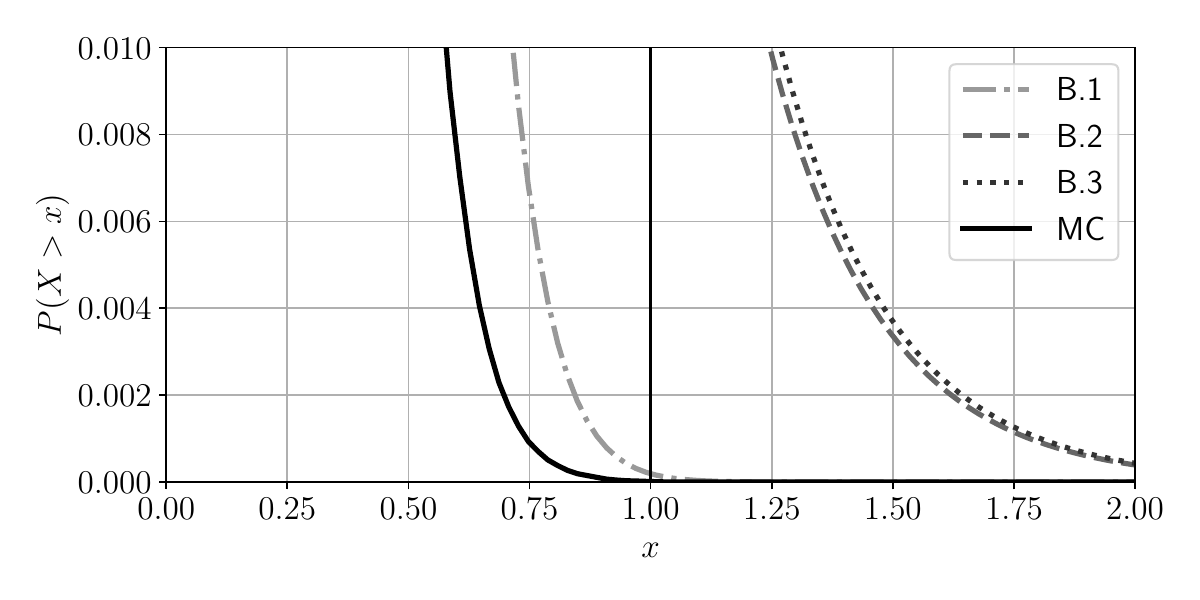}
    \end{center}
    \caption{%
        The diagram considers the toy GP from Section~\ref{subsection_toy_example}, see also Figure~\ref{figure_centered_GP} for a visualization of this GP.
        It shows the tail distribution (complementary cumulative distribution function) $P(X > x)$ for different values of $x$ of different estimations for the supremum of the centered GP. Note, that the safety condition here is $P(X < 1)$
        The MC bound can be seen as optimal estimation of $P(X>x)$.
        Amongst the upper bounds to $P(X>x)$, we see, that the strong Borell inequality \ref{eqn:Borell-1} is the sharpest one.
    }
	\label{figure_toy_comparison}
\end{figure*}
We call this method Adaptive Borell (AB).
A general adaptive algorithm is provided in Algorithm~\ref{alg:adaptive}, while method specific versions are in the supplement.
\begin{algorithm}[t]
\caption{Adaptive Safety evaluation}
    \begin{algorithmic}
        \Require Safety threshold: $\alpha>0$,\\
        Threshold for confidence intervals: $\epsilon>0$,\\
        Discretization: $t_1,\dots,t_m$,\\
        Sample sizes: $0=M_0 < M_1 < \ldots < M_R$, \\
        Posterior GP: $X_t$.
        \State $\widehat{P} \gets 0$
        \For{$r=1,\ldots, R$}%
            \For{$i=M_{r-1}+1,\ldots, M_r$}
                \State Simulate $(X_{t_j,i})_{j=1,\ldots, m}$
                \State $S_i \gets   \max_{j=1,\ldots, m} X_{t_j,i}$
            \EndFor
            \State Update $\widehat{P}$ given $S_i$
            \State Compute confidence intervals $[\widehat{P}_-, \widehat{P}_+]$ 
            \If{$\widehat{P}_+ \leq \alpha$} 
                \State \Return \textsc{safe}
            \ElsIf{$\widehat{P}_- \geq \alpha$}
                \State \Return \textsc{unsafe}
            \EndIf
        \EndFor
        \State \Return \textsc{unsafe}
    \end{algorithmic}
    \label{alg:adaptive}
\end{algorithm}

\subsection{Hybrid adaptive sampling (ABM)}

If the assessed trajectory is either clearly safe ($P^*\ll \alpha$) or clearly unsafe ($P^*\gg \alpha$), the adaptive MC scheme might terminate earlier than the semi-analytical procedure.
As both methods use the same samples, there is minimal computational overhead to run both schemes in parallel, with remaining uncertainty $\epsilon/2$ instead of $\epsilon$, and stop as soon as one of them reaches a decision.

Moreover, AB provides an upper bound on the unsafeness probability, $P^\dagger \geq P^*$. 
That is, even if $P^\dagger>\alpha$ (AB bound classifies a trajectory as unsafe), we might still have $\alpha\geq P^*$ (trajectory is indeed safe). 
Thus, the AB method is overly confident, which maintains safety, but hinders exploration.
As a further improvement, we suggest to run both adaptive methods in parallel, and use the Borell-TIS bound only to conclude safety, but not unsafety.
More precisely, we stop sampling at step
\begin{align*}
    r^\diamond=\inf \Big\{r\;:\; 
	   & \quad \widehat{P}^\dagger_{+}(M_r,r, \tfrac{\epsilon}{2}) < \alpha \\ & \;\text{or}\; \widehat{P}_{\textsc{MC}}^+(\tau, M_r, \tfrac{\epsilon}{2}, \alpha) < \alpha \\ &\;\text{or}\;  \widehat{P}_{\textsc{MC}}^-(\tau, M_r, \epsilon, \alpha) > \alpha) \Big\}.
\end{align*}
In the notation of Algorithm \ref{alg:adaptive}, we have a sequential confidence interval given by the bounds
\begin{align*}
    \widehat{P}^\diamond_+(r,\epsilon,\alpha) &:= \min\left[\widehat{P}^\dagger_{+}(M_r,r, \tfrac{\epsilon}{2}),\, \widehat{P}_{\textsc{MC}}^+(\tau, M_r, \tfrac{\epsilon}{2},\alpha)\right], \\
     \widehat{P}^\diamond_-(r,\epsilon,\alpha) &:= \widehat{P}_{\textsc{MC}}^-(\tau, M_r, \epsilon,\alpha).
\end{align*}

We call this method Adaptive Borell-Monte-Carlo (ABM).
Theorem \ref{thm:adaptive} and Theorem \ref{thm:adaptive-Borell} directly imply that the proposed hybrid scheme makes no wrong decisions about safety of a trajectory, with probability $1-\epsilon$, which proves the following corollary.

\begin{corollary}\label{coro:adaptive-combined}
	Let $Q$ denote the probability w.r.t.\ the MC sampling and let $\epsilon\in(0,1)$.
    If ${P}^*(\tau)\geq \alpha$ then
	\begin{align*}
		Q\left( \exists r\in\mathbb{N}: \, \widehat{P}^\diamond_+(r,\epsilon,\alpha) < \alpha \right) \leq \epsilon
	\end{align*}
	and if ${P^*}(\tau)\leq \alpha$, then
	\begin{align*}
		Q\left( \exists r\in\mathbb{N}: \, \widehat{P}^\diamond_-(r,\epsilon,\alpha) > \alpha \right) \leq \epsilon.
	\end{align*}
\end{corollary}

\begin{figure*}[ht]
    \begin{center}
        \input{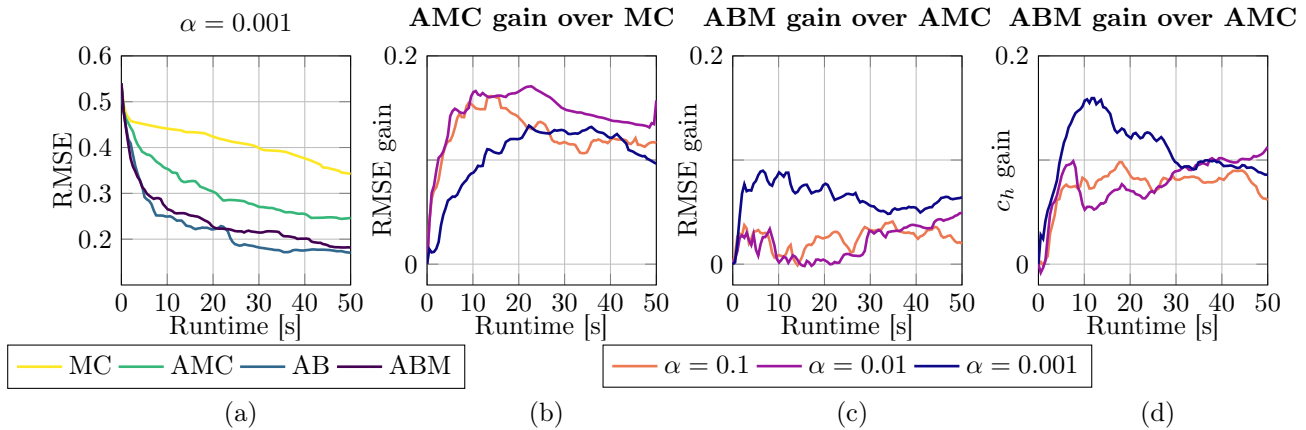}
    \end{center}
    \caption{
        Consider the Himmelblau's function exploration from Section~\ref{subsection_himmelblau}.
        (a) All our three novel adaptive methods (AMC, AB, ABM) improve upon the current state of the art MC with a sufficient sample size (see supplement for exact numbers), when considering the RMSE for high safety requirements {\color{alphacolor3}$\alpha=0.001$}.
        (b) Our method AMC improves over MC consistently in RMSE for three different safety requirements.
        Our favoured method ABM improves further over AMC both for (c) RMSE and (d) detecting the safe region correctly via the health coverage $c_h$.
        Results are averaged over 10 independent seeds.}
	\label{figure_himmelblau_comparison}
\end{figure*}

\section{EXAMPLES}\label{section_examples}

In this section, we test our method using simulated experiments. For multidimensional examples we use linear ramps with equidistant points as trajectories, similar to \cite{zimmer2018safe}. 
In our experiments, the computational time budget for each method is fixed, and the number of measurements $n_{\text{SAL}}$ which can be taken during one run should be as high as possible.
Moreover, we consider different performance metrics to compare the classic Monte-Carlo (MC) method to our novel adaptive Monte-Carlo (AMC) method, adaptive Borell (AB) method, and adaptive Borell-Monte-Carlo (ABM) method. \\
The root-mean-squared error (RMSE) between the obtained GP model and the ground truth indicates the quality of actively learning the behaviour of the real world system.
As a second performance metric, we consider the health coverage, defined as the accuracy of a binary classifier which uses the posterior mean $\mu(x)$ to classify the domain as safe or unsafe, i.e.\
\begin{equation*}
    c_h = \frac{\int_\mathcal{X} \mathds{1}_{\mu(x) \geq 0} \mathds{1}_{z \geq 0}dx + \int_\mathcal{X} \mathds{1}_{\mu(x) < 0} \mathds{1}_{z < 0}dx}{\int_\mathcal{X} 1 dx}.
\end{equation*}
For computational tractability, the integrals are approximated via a fixed number of discrete evaluations.

\subsection{Univariate Toy Example}\label{subsection_toy_example}

First, we omit the active learning and only assess the quality of our safety evaluation.
To this end, we consider the toy example $f \colon [0, 1] \to \R, x \mapsto -0.2\sin(10x) - x + 1.1$ with a GP generated by the squared exponential kernel and hyperparameters $\sigma_f = 1$, $\ell^2 = 32^{-1}$ and $\sigma_N^2 = 10^{-3}$. 
The training points $0= x_1\le x_2\le \ldots\le x_{21}= 1$ are equally spaced, and we consider the safety of the posterior trajectory on the domain $[0,1]$.
For a visualization of this GP, see Figure \ref{figure_centered_GP}.
We compare the bounds from our methods with what we regard as the true tail probability: the bound generated by MC sampling with a high number of samples ($M=10^6)$. 
We discretize the posterior trajectories by 50 equidistant points, i.e.\ we approximate $P_{\text{unsafe}}\approx P^*$ with $T=\{0, \frac{1}{49},\ldots, 1\}$
The comparison is depicted in Figure \ref{figure_toy_comparison}.
The bound \ref{eqn:Borell-1} using the median performs best and is really close to the MC simulation. 
The bounds given by \ref{eqn:Borell-2} and \ref{eqn:Borell-3} behave poorly in the tails, as the subgaussian tails are fatter than the Gaussian tails.
These numerical results indicate that the bound of AB derived from \ref{eqn:Borell-1} is rather tight.

\subsection{Himmelblau's function exploration}
\label{subsection_himmelblau}

As an example of an active learning task, we consider exploration of a version of Himmelblau's function $f(x, y) = (x^2+y-11)^2 + (x+y^2-7)^2$ \citep{himmelblau2018applied}. 
In particular, we want to actively learn the function $f$ in the region $[-3, 3]^2$, with the safety constraint $f(x,y) \geq 50$, for a visualization, see Figure \ref{fig:contribution}.
Thus, we have a connected safe area with the unsafe area only at the boundary of the given square.
For further details on the GP setting, see the supplementary.

The results are shown in Figure \ref{figure_himmelblau_comparison}. 
The experiments support the theoretical claims, that adaptive methods perform better with rising safety requirements, see Figure \ref{figure_himmelblau_comparison} (a). This is based on a fewer runtime per iteration due to less MC samples needed to make a provably right decision with high possibility. Indeed, the methods containing \ref{eqn:Borell-1} use to make up to 6.5 times as many iterations in the same time, see Figure \ref{fig:contribution}. 

\subsection{Application: Engine control}\label{subsection_railpressure}
\begin{figure}[t]
    \centering
    \includegraphics[scale=0.36]{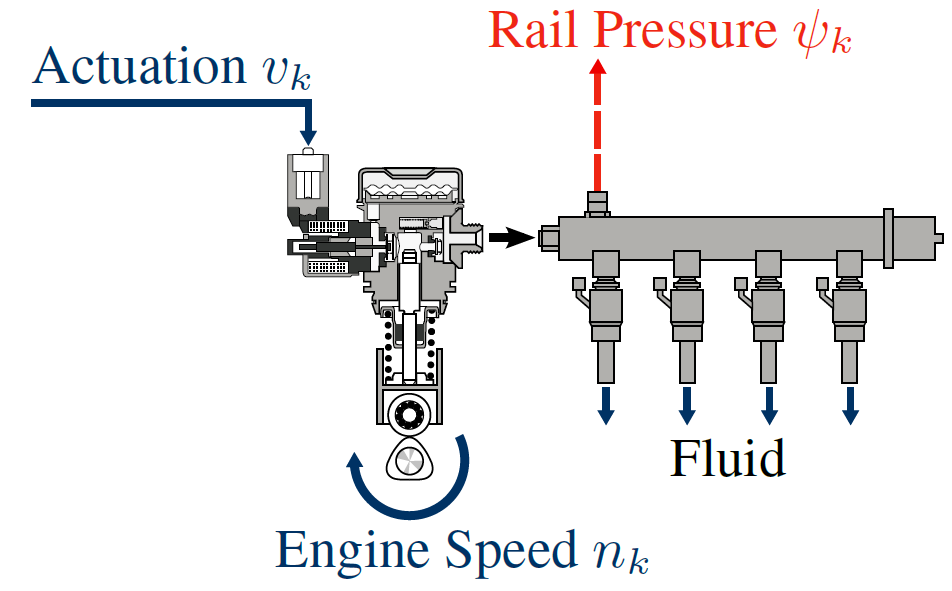}
    \caption{High-pressure fluid injection system with controllable inputs $v_k, n_k$ and measured output $\psi_k$ (picture taken from \cite{zimmer2018safe,tietze2014model})}
    \label{fig:Railpressure}
\end{figure}

\begin{figure*}[ht]
    \begin{center}
        \input{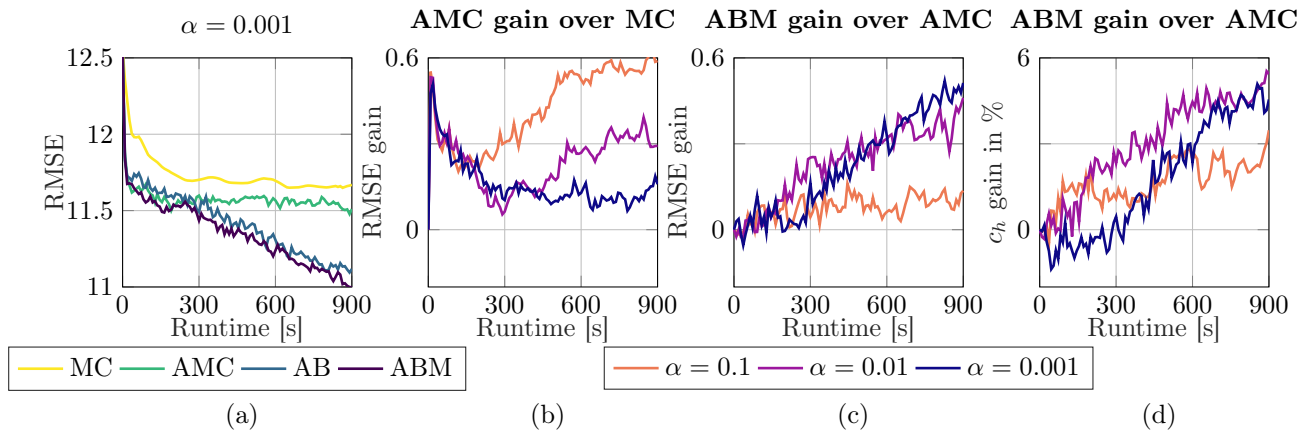}
    \end{center}
    \caption{%
        Consider the engine control exploration from Section~\ref{subsection_railpressure}.
        (a) Both our methods AB and ABM using the Borell-TIS inequality \ref{eqn:Borell-1} improve upon the current state of the art MC and its adaptive version AMC for high safety requirements {\color{alphacolor3}$\alpha=0.001$}.
        (b) The adaptive version AMC improves over MC in RMSE in particular for the lax safety requirement {\color{alphacolor1}$\alpha=0.1$}.
        (c) For higher safety requirements {\color{alphacolor2}$\alpha=0.01$} and {\color{alphacolor3}$\alpha=0.001$} our favoured method ABM improves over AMC and hence also over MC in RMSE.
        (d) The enhanced RMSE results observed in (c) can be primarily attributed to the improved health coverage $c_h$.
        Results are averaged over 10 independent seeds.
    }
	\label{figure_rp_comparison}
\end{figure*}

We consider a dynamic high-pressure fluid system for fuel injection in combustion engines introduced in \cite{zimmer2018safe} as a real world example, see Figure \ref{fig:Railpressure}.
Actuation $v_k$ and the engines' speed $n_k$ are inputs every time step $k$, and we aim to learn a surrogate model for the rail pressure $\psi_k$.
We inherit the assumption of a nonlinear autoregressive structure $\psi_k=\psi(x_k)$ for $x_k = (n_k, n_{k-1}, n_{k-2}, n_{k-3}, v_k, v_{k-1}, v_{k-3})$. 
The trajectories linearly interpolate two points equidistantly. For more details about the GP settings, see supplementary.
The results are depicted in Figure \ref{figure_rp_comparison}, in analogy to Figure \ref{figure_himmelblau_comparison}. 
Again, our adaptive methods show improved performance over standard MC, see Figure \ref{figure_rp_comparison} (a). Especially for higher safety requirements, the methods containing \ref{eqn:Borell-1} outperform the others, see Figure \ref{figure_rp_comparison} (a),(c) and (d). With fewer needed MC samples, AB and ABM can perform more iterations in the same runtime, see the supplementary for details.

\section{CONCLUSION}

In this paper, we explore the derivation of upper bounds on the probabilities of sampled Gaussian processes exceeding prescribed thresholds.
Leveraging adaptive techniques, we achieve significant computational efficiency enhancements compared to state-of-the-art Monte-Carlo sampling methods.
Furthermore, we extend these advancements by incorporating a variant of the Borell-TIS inequality in conjunction with classical Monte-Carlo sampling.
While the Borell-TIS inequality itself entails sampling, it serves to estimate the median rather than the direct estimation of tail probabilities.
To facilitate the application of the Borell-TIS inequality, we introduce a centering transformation for Gaussian processes and offer an insightful interpretation.
We rigorously establish error bounds for all of our probabilistic methods, ensuring their reliability in practical applications.

Our primary motivation revolves around the domain of safe active learning in dynamic systems.
Although our Gaussian process bounds tend to be conservative, potentially resulting in slower exploration, the remarkable reduction in computation time allows for the acquisition of more data points when employing our methods.
This advantageous trade-off effectively offsets the conservative nature of our bounds, as empirically demonstrated in our illustrative examples.
Notably, our approach proves particularly valuable in scenarios with stringent safety requirements, such as when the trajectory safety probability must exceed 99.9\%.

\section*{Acknowledgements}
Jörn Tebbe is supported by the SAIL project which is funded by the Ministry of Culture and Science of the State of North Rhine-Westphalia under the grant no NW21-059C. \\
Fabian Mies was employed at RWTH Aachen University while this research was conducted.

\renewcommand\refname{\vspace{-1.2em}} %
\bibliography{literature}

\begin{thebibliography}{}

\bibitem[Adler and Taylor, 2007]{Adler2007}
Adler, R.~J. and Taylor, J.~E. (2007).
\newblock {\em Random {{Fields}} and {{Geometry}}}.
\newblock {Springer New York}.

\bibitem[Baumann et~al., 2021]{Baumann21}
Baumann, D., Marco, A., Turchetta, M., and Trimpe, S. (2021).
\newblock Gosafe: Globally optimal safe robot learning.
\newblock {\em ICRA}.

\bibitem[Berkenkamp et~al., 2016]{Berkenkamp16}
Berkenkamp, F., Schoellig, A.~P., and Krause, A. (2016).
\newblock Safe controller optimization for quadrotors with gaussian processes.
\newblock {\em ICRA}.

\bibitem[Besginow and Lange-Hegermann, 2022]{besginow2022constraining}
Besginow, A. and Lange-Hegermann, M. (2022).
\newblock Constraining {G}aussian processes to systems of linear ordinary
  differential equations.
\newblock {\em NeurIPS}.

\bibitem[Bitzer et~al., 2022]{Bitzer2022}
Bitzer, M., Meister, M., and Zimmer, C. (2022).
\newblock Structural kernel search via bayesian optimization and symbolical
  optimal transport.
\newblock {\em NeurIPS}.

\bibitem[Borovitskiy et~al., 2020]{borovitskiy2020matern}
Borovitskiy, V., Terenin, A., Mostowsky, P., et~al. (2020).
\newblock Mat{\'e}rn {G}aussian processes on riemannian manifolds.
\newblock {\em NeurIPS}.

\bibitem[Botev, 2017]{botev2017normal}
Botev, Z.~I. (2017).
\newblock The normal law under linear restrictions: simulation and estimation
  via minimax tilting.
\newblock {\em Journal of the Royal Statistical Society Series B: Statistical
  Methodology}, 79(1):125--148.

\bibitem[Bottero et~al., 2022]{bottero2022information}
Bottero, A., Luis, C., Vinogradska, J., Berkenkamp, F., and Peters, J.~R.
  (2022).
\newblock Information-theoretic safe exploration with {G}aussian processes.
\newblock {\em NeurIPS}.

\bibitem[Cardelli et~al., 2019]{cardelli2019robustness}
Cardelli, L., Kwiatkowska, M., Laurenti, L., and Patane, A. (2019).
\newblock Robustness guarantees for bayesian inference with {G}aussian
  processes.
\newblock {\em AAAI}.

\bibitem[Conover, 1999]{conover1999}
Conover, W.~J. (1999).
\newblock {\em Practical Nonparametric Statistics}.
\newblock Probability and Statistics: {{Applied}} Probability and Statistics
  Section. Wiley.

\bibitem[Dudley, 1967]{dudley1967sizes}
Dudley, R.~M. (1967).
\newblock The sizes of compact subsets of hilbert space and continuity of
  gaussian processes.
\newblock {\em Journal of Functional Analysis}, 1(3):290--330.

\bibitem[Duvenaud et~al., 2013]{duvenaud2013structure}
Duvenaud, D., Lloyd, J., Grosse, R., Tenenbaum, J., and Zoubin, G. (2013).
\newblock Structure discovery in nonparametric regression through compositional
  kernel search.
\newblock {\em ICML}.

\bibitem[Duvenaud et~al., 2011]{duvenaud2011additive}
Duvenaud, D.~K., Nickisch, H., and Rasmussen, C. (2011).
\newblock Additive {G}aussian processes.
\newblock {\em NeurIPS}.

\bibitem[Gardner et~al., 2018]{gardner2018gpytorch}
Gardner, J., Pleiss, G., Weinberger, K.~Q., Bindel, D., and Wilson, A.~G.
  (2018).
\newblock {GPytorch: Blackbox matrix-matrix {G}aussian process inference with
  GPU acceleration}.
\newblock {\em NeurIPS}.

\bibitem[Genz, 1992]{genz1992numerical}
Genz, A. (1992).
\newblock Numerical computation of multivariate normal probabilities.
\newblock {\em Journal of computational and graphical statistics},
  1(2):141--149.

\bibitem[Gessner et~al., 2020]{gessner2020integrals}
Gessner, A., Kanjilal, O., and Hennig, P. (2020).
\newblock Integrals over gaussians under linear domain constraints.
\newblock In {\em International conference on artificial intelligence and
  statistics}, pages 2764--2774. PMLR.

\bibitem[H{\"a}rk{\"o}nen et~al., 2023]{harkonen2022gaussian}
H{\"a}rk{\"o}nen, M., Lange-Hegermann, M., and Rai{\c{t}}{\u{a}}, B. (2023).
\newblock {G}aussian process priors for systems of linear partial differential
  equations with constant coefficients.
\newblock {\em ICML}.

\bibitem[Hensman et~al., 2017]{hensman2017variational}
Hensman, J., Durrande, N., Solin, A., et~al. (2017).
\newblock Variational {F}ourier features for {G}aussian processes.
\newblock {\em JMLR}.

\bibitem[Hensman et~al., 2013]{hensman2013gaussian}
Hensman, J., Fusi, N., and Lawrence, N.~D. (2013).
\newblock {G}aussian processes for big data.
\newblock {\em UAI}.

\bibitem[Himmelblau et~al., 2018]{himmelblau2018applied}
Himmelblau, D.~M. et~al. (2018).
\newblock {\em Applied nonlinear programming}.
\newblock McGraw-Hill.

\bibitem[Holderrieth et~al., 2021]{holderrieth2021equivariant}
Holderrieth, P., Hutchinson, M.~J., and Teh, Y.~W. (2021).
\newblock Equivariant learning of stochastic fields: {G}aussian processes and
  steerable conditional neural processes.
\newblock {\em ICML}.

\bibitem[L{{\'a}}zaro-Gredilla et~al., 2010]{lazaro20sparse}
L{{\'a}}zaro-Gredilla, M., Qui{{\~n}}nero-Candela, J., Rasmussen, C.~E., and
  Figueiras-Vidal, A.~R. (2010).
\newblock Sparse spectrum {G}aussian process regression.
\newblock {\em JMLR}.

\bibitem[Lederer et~al., 2019]{lederer2019}
Lederer, A., Umlauft, J., and Hirche, S. (2019).
\newblock Uniform {Error} and {Posterior} {Variance} {Bounds} for {Gaussian}
  {Process} {Regression} with {Application} to {Safe} {Control}.
\newblock {\em NeurIPS}.

\bibitem[Li et~al., 2022]{Li2022}
Li, C.-Y., Rakitsch, B., and Zimmer, C. (2022).
\newblock Safe active learning for multi-output gaussian processes.
\newblock {\em AISTATS}.

\bibitem[Rasmussen et~al., 2006]{rasmussen2006gaussian}
Rasmussen, C.~E., Williams, C.~K., et~al. (2006).
\newblock {\em {G}aussian processes for machine learning}.
\newblock MIT Press.

\bibitem[Sandmeier, 2022]{sandmeier2022optimization}
Sandmeier, N. (2022).
\newblock {\em Optimization of adaptive test design methods for the
  determination of steady-state data-driven models in terms of combustion
  engine calibration}.
\newblock Universit{\"a}tsverlag der Technischen Universit{\"a}t Berlin.

\bibitem[Schreiter et~al., 2015]{schreiter2015safe}
Schreiter, J., Nguyen-Tuong, D., Eberts, M., Bischoff, B., Markert, H., and
  Toussaint, M. (2015).
\newblock Safe exploration for active learning with {G}aussian processes.
\newblock {\em ECML PKDD}.

\bibitem[Settles, 2009]{Settles09}
Settles, B. (2009).
\newblock Active learning literature survey.
\newblock {\em Computer Sciences Technical Report 1648}.

\bibitem[Sui et~al., 2018]{sui2018stagewise}
Sui, Y., Zhuang, V., Burdick, J., and Yue, Y. (2018).
\newblock Stagewise safe bayesian optimization with gaussian processes.
\newblock {\em ICML}.

\bibitem[Tharwat and Schenck, 2023]{tharwat2023survey}
Tharwat, A. and Schenck, W. (2023).
\newblock A survey on active learning: State-of-the-art, practical challenges
  and research directions.
\newblock {\em Mathematics}, 11(4):820.

\bibitem[Thewes et~al., 2016]{thewes2016efficient}
Thewes, S., Krause, M., Reuber, C., Lange-Hegermann, M., Dziadek, R., and
  Rebbert, M. (2016).
\newblock Efficient in-vehicle calibration by the usage of automation and
  enhanced online doe approaches.
\newblock {\em Simulation and Testing for Vehicle Technology: 7th Conference}.

\bibitem[Tietze et~al., 2014]{tietze2014model}
Tietze, N., Konigorski, U., Fleck, C., and Nguyen-Tuong, D. (2014).
\newblock Model-based calibration of engine controller using automated
  transient design of experiment.
\newblock {\em 14. Internationales Stuttgarter Symposium: Automobil-und
  Motorentechnik}.

\bibitem[Titsias, 2009]{titsias2009variational}
Titsias, M. (2009).
\newblock Variational learning of inducing variables in sparse {G}aussian
  processes.
\newblock {\em AISTATS}.

\bibitem[van~der Vaart and Wellner, 1996]{VanderVaart1996}
van~der Vaart, A.~W. and Wellner, J.~A. (1996).
\newblock {\em Weak {Convergence} and {Empirical} {Processes}}.
\newblock Springer New York.

\bibitem[Wang et~al., 2019]{wang2019exact}
Wang, K., Pleiss, G., Gardner, J., Tyree, S., Weinberger, K.~Q., and Wilson,
  A.~G. (2019).
\newblock Exact {G}aussian processes on a million data points.
\newblock {\em NeurIPS}.

\bibitem[Wilson and Nickisch, 2015]{wilson2015kernel}
Wilson, A. and Nickisch, H. (2015).
\newblock {Kernel interpolation for scalable structured {G}aussian processes
  (KISS-GP)}.
\newblock {\em ICML}.

\bibitem[Zimmer et~al., 2020]{zimmer2020}
Zimmer, C., Driess, D., Meister, M., and Duy, N.-T. (2020).
\newblock Adaptive discretization for evaluation of probabilistic cost
  functions.
\newblock {\em AISTATS}.

\bibitem[Zimmer et~al., 2018]{zimmer2018safe}
Zimmer, C., Meister, M., and Nguyen-Tuong, D. (2018).
\newblock Safe active learning for time-series modeling with {G}aussian
  processes.
\newblock {\em NeurIPS}.

\end{thebibliography}
 
 \onecolumn
\aistatstitle{Efficiently Computable Safety Bounds for Gaussian Processes in
Active Learning \\
Supplementary Materials}
\section{Discrete and continuous trajectories}
The theory and methodology developed in this paper is formulated in terms of continuous-time paths of Gaussian processes $Z_t$, parameterized by some trajectory $\tau(t)\in\mathbb{R}^n$, for $t\in T\subset [0,1]$. 
The set $T$ is introduced to account for a discretization of the trajectory.
Here, we highlight that we could instead describe the discretization by keeping $T=[0,1]$ and considering the alternative trajectory 
\begin{align*}
    \tau(t) = \begin{cases}
        \tau_1, & t\in [0, t_2), \\
        \tau_i, & t\in [t_i, t_{i+1}),\quad i=1,\ldots, m-1 \\
        \tau_m, & t\in [t_m, 1].
    \end{cases}
\end{align*}
Here, $\tau_i\in \mathbb{R}^n$ are the discrete points of the trajectory, and $0\leq t_1\leq \tau_2 \leq \ldots \leq \tau_m\leq 1$ are breakpoints of the step function. As $Z_t$ is the posterior GP of $f(\tau(t))$, its paths are also step functions, and it holds that
\begin{align*}
    \inf_{t\in [0,1]} Z_t = \min_{j=1,\ldots, m} Z_{t_j}, \qquad   \text{and} \qquad P^*(\tau) = P\left( \min_{j=1,\ldots, m} Z_{t_j} \leq 0\right).
\end{align*}
Thus, all of our results readily transfer to finite discretizations of the trajectories $\tau$.
In particular, our methods are applicable for the discretizations chosen in Section 5.
\section{Proofs}
\subsection{Proof of Theorem 1}
	Observe that $M \cdot \widehat{P}_{\textsc{MC}}(M,\tau)$ admits a Binomial distribution with $M$ trials and success probability $p = P^*(\tau)$.
    Moreover, let $Y_r$ be a binomially distributed random variable with $M_r$ trials and success probability $\alpha$, and set $X=Y_r/M_r$.
    First, consider the case $p\geq \alpha$, such that in particular $X_r$ is stochastically smaller than $\widehat{P}_{\textsc{MC}}(M_r,\tau)$. 
    We now use Okamoto's exponential bounds \cite[Thm.~2]{okamoto1959} for the Binomial distribution: As $ \alpha \le 1/2 $, for each $ z > 0 $ it holds
 $ Q(X_r - \alpha \le -z ) \le \exp( - \frac{M_r z^2}{2 \alpha(1-\alpha)} ) $.
    Thus, the union bound yields
    
   \begin{align*}
        Q\left(\exists r \in \mathbb{N} : \widehat{P}^{+}_{\textsc{MC}}(\tau,M_r,r,\epsilon,\alpha) < \alpha \right) 
       &\leq \sum_{r=1}^\infty Q\left(\widehat{P}_{\textsc{MC}}(M_r, \tau) < \alpha - \sqrt{\alpha(1-\alpha)} c_r \right) \\
       &\leq \sum_{r=1}^\infty Q\left(X_r - \alpha < - \sqrt{\alpha(1-\alpha)} c_r \right) \\
       &\le \sum_{r=1}^\infty \exp\left( - \frac{M_r c_r^2}{2}\right) \\
       & = \sum_{r=1}^\infty \frac{6\epsilon}{\pi^2 r^2} = \epsilon.
   \end{align*}
   \newpage
   For the case $p\leq \alpha\leq \frac{1}{2}$, the random variable $X_r$ is stochastically larger than $\widehat{P}_{\textsc{MC}}(M_r,\tau)$.
   Hence, we obtain
   \begin{align*}
        Q\left(\exists r \in \mathbb{N} : \widehat{P}^{-}_{\textsc{MC}}(\tau,M_r,r,\epsilon,\alpha) > \alpha \right) 
        &\leq  \sum_{r=1}^\infty Q\left(\widehat{P}_{\textsc{MC}}(M_r, \tau) > \alpha + \frac{c_r^2}{4} + c_r\sqrt{\alpha} \right) \\
        &=  \sum_{r=1}^\infty Q\left(\widehat{P}_{\textsc{MC}}(M_r, \tau) > \left(\sqrt{\alpha} + \frac{c_r}{2}\right)^2 \right) \\
        &\leq  \sum_{r=1}^\infty Q\left( X_r > \left(\sqrt{\alpha} + \frac{c_r}{2}\right)^2 \right).
   \end{align*}
   Now we use another bound of Okamoto \cite[Thm.~3]{okamoto1959}: $Q( X_r > (\sqrt{\alpha}+z)^2 ) \leq \exp(-2M_rz^2)$.
    Hence,
    \begin{align*}
     Q\left(\exists r \in \mathbb{N} : \widehat{P}^{-}_{\textsc{MC}}(\tau,M_r,r,\epsilon,\alpha) > \alpha \right)
     &\leq \sum_{r=1}^\infty Q\left( X_r > \left(\sqrt{\alpha} + \frac{c_r}{2}\right)^2 \right) \\
         &\leq \sum_{r=1}^\infty\exp\left( -\frac{M_r c_r^2}{2} \right) \\
         &= \sum_{r=1}^\infty \frac{6\epsilon}{\pi^2r^2} \quad = \epsilon.
    \end{align*}
    This completes the proof.
 
\subsection{Proof of Theorem 5}
	Suppose that $P^\dagger(\tau)\geq \alpha$.
	Our choice of $\beta_+$ ensures that
    \begin{align*}
        Q\left( \tilde{m}\leq q_{\beta_+, M_r} \right) &\geq \chi(r,\epsilon) = 1-\frac{6\epsilon}{\pi^2r^2} \\
        \qquad \iff Q\left( \tilde{m}> q_{\beta_+, M_r} \right) &\leq \frac{6\epsilon}{\pi^2r^2}.
    \end{align*}
    Hence,
	\begin{align*}
		Q\left( \widehat{P}^\dagger_{+}(M_r,r,\epsilon)< \alpha \right)
        &= Q\left( 1-\Phi\left( \tfrac{1-q_{\beta_+, M_r}}{\tilde{\sigma}} \right) < \alpha \leq P^\dagger(\tau) \right)  \\
        &\leq Q\left( 1-\Phi\left( \tfrac{1-q_{\beta_+, M_r}}{\tilde{\sigma}} \right) <  P^\dagger(\tau) \right)  \\
        &= Q\left( 1-\Phi\left( \tfrac{1-q_{\beta_+, M_r}}{\tilde{\sigma}} \right) <  1-\Phi\left( \tfrac{1-\tilde{m}}{\tilde{\sigma}} \right) \right)  \\
        &= Q\left( q_{\beta_+, M_r} <  \tilde{m}\right) \\
        &\leq \frac{6\epsilon}{\pi^2 r^2}.
	\end{align*}
	The union bound yields
	\begin{align*}
		Q\left( \exists r\in\mathbb{N}: \, \widehat{P}^\dagger_{+}(M_r,r,\epsilon)< \alpha \right) 
		\;\leq\; \sum_{r=1}^\infty Q\left( \widehat{P}^\dagger_{+}( M_r,r,\epsilon)< \alpha \right) 
		\;\leq\; \sum_{r=1}^\infty \frac{6\epsilon}{\pi^2 r^2}
		\;\leq\; \epsilon.
	\end{align*}
	This proves the first claim, and the second claim may be derived analogously. %

\subsection{Proof of Corollary 6}
    Suppose that $P^*(\tau) \geq \alpha$. 
    By the definition of $\widehat{P}_+^\diamond(r,\epsilon, \alpha)$ and the union bound, we find that
    \begin{align*}
        Q\left( \exists r\in\mathbb{N}:\, \widehat{P}_+^\diamond(r,\epsilon, \alpha) <\alpha  \right)  
        & \leq Q\left( \exists r\in\mathbb{N}:\, \widehat{P}_+^\dagger(M_r, r, \tfrac{\epsilon}{2}) <\alpha  \right) \\
        &\quad + Q\left( \exists r\in\mathbb{N}:\, \widehat{P}^{+}_{\textsc{MC}}(\tau, M_r, r, \tfrac{\epsilon}{2},\alpha) <\alpha  \right) \\
        &\leq \tfrac{\epsilon}{2} +\tfrac{\epsilon}{2}. 
    \end{align*}
    The last inequality is due to Theorem 1 and Theorem 5, and establishes the first claim of Corollary 6. The second claim is identical to Theorem 1.

\subsection{Further Proofs}
The proof of Theorem 2 can be found in A.2.1 in \cite{VanderVaart1996}.
The proof of Theorem 4 is given in the main text as a combination of Theorem 2 and Remark 3.

\section{Code}
The code is provided under \url{github.com/joerntebbe/SafetyBounds4GPinAL} with the BSD-2 license. We use adapted code from \cite{zimmer2018safe}, which is provided under the MIT license, with a modified version of the Gaussian Process library from \cite{rasmussen2010gaussian} for MATLAB, which is provided under the FreeBSD license. \\
The experiments were carried out using CPU only with an AMD Ryzen 9 5950X driven at 3.4GhZ and 64GB RAM on MATLAB R2023a.

\section{Algorithms}
\subsection{General implementation details}

We use the active learning algorithm proposed in \cite{zimmer2018safe} as baseline. 
Integrating our method into this algorithm comes with several challenges. \\
We have to add an immediate rejection of a candidate trajectory, if the mean has a changing sign, since our method is not applicable in this case. In order to provide meaningful gradients to the optimizer in this case, we implemented a heuristic to return a safety value which characterizes the trajectory as unsafe but also gives a metric on how far away the trajectory is from being safe without evaluating a particular bound. This results in a penalty term which is dependent on the distance from the mean to be classifiable by our method

\begin{equation*}
    {P}_{\text{unsafe}}(\tau_t) = 0.5 + \Vert \max(\mu_t, \textbf{0}_m) \Vert_2
\end{equation*}
with $\textbf{0}_m \in \mathbb{R}^m$ being a vector of zeros. \\
Moreover, if our method fails to make a decision with the provided budget of samples per safety evaluation, we declare the trajectory as unsafe. With the same purpose as before, we want to provide a numerical value which yields how far away from safe this trajectory is. In order to do so, we provide the lower bound of the confidence interval as the returning value.

As the sample sizes $M_1, \dots ,M_R$ we use $M_1 = 100$ and double the size in each iteration, resulting in \[M_r = 100 \cdot 2^{r-1}.\] For the Himmelblau example we use $R=14$, for the engine control example we use $R=17$. These values were chosen based on prior experiments which observed the distribution of required samples to make a decision. In order to provide a fair comparison, i.e. the MC method respects the safety requirements, we chose the fixed sample size for the MC method to be $M_{R-1}$. 
\newpage
\subsection{Adaptive Monte Carlo}
Algorithm \ref{alg:adaptive-MC} provides Pseudocode for the adaptive Monte Carlo sampling scheme.
\begin{algorithm}[H]
\caption{Adaptive Monte Carlo sampling}
    \begin{algorithmic}
        \Require Safety threshold: $\alpha>0$, Threshold for confidence intervals $\epsilon>0$, \\
        Discretization $t_1, \dots, t_m$, sequence of sample sizes $0=M_0 < M_1 < \ldots < M_R$,
        Posterior GP: $X_t$ 
        \For{$r=1,\ldots, R$}%
            \For{$i=M_{r-1}+1,\ldots, M_r$}
                \State Simulate $(X_{t_j,i})_{j=1,\ldots, m}$
                \State $S_i \gets   \max_{j=1,\ldots, m} X_{t_j,i} $
            \EndFor
            \State $\widehat{P} \gets \frac{M_{r-1}}{M_r} \widehat{P} + \frac{1}{M_r} \sum_{i=M_{r-1}+1}^{M_r} \mathbf{1}\left(S_i>1\right)$
            \State $\widehat{P}_{\text{MC}}^\pm \gets \widehat{P} \pm \sqrt{ \frac{2\alpha(1-\alpha)}{M_r} \left| \log \frac{6\epsilon}{\pi^2 r^2} \right| }$
            \If{$\widehat{P}_{\text{MC}}^+ <\alpha$} 
                \State \Return \textsc{safe}
            \ElsIf{$\widehat{P}_{\text{MC}}^- >\alpha$}
                \State \Return \textsc{unsafe}
            \EndIf
        \EndFor
        \State \Return \textsc{unsafe}
    \end{algorithmic}
    \label{alg:adaptive-MC}
\end{algorithm}
\subsection{Adaptive Borell-TIS}

Algorithm \ref{alg:cap} provides Pseudocode for the safety evaluation using the Borell-TIS inequality and the adaptive sampling scheme of the median.

\begin{algorithm}[h]
\caption{Adaptive sampling for Borell-TIS}\label{alg:cap}
    \begin{algorithmic}
        \Require Safety threshold: $\alpha>0$, Threshold for confidence intervals $\epsilon>0$, \\
        Discretization $t_1, \dots, t_m$, sequence of sample sizes $0=M_0 < M_1 < \ldots < M_R$,
        Posterior GP: $X_t$  
        \For{$r=1,\ldots, R$}
                \State $\beta_\pm \gets \frac{1}{2} \pm \Phi^{-1}(1-\chi(M_r,\epsilon))/\sqrt{4M_r}$ %
                \For{$i=M_{r-1}+1,\ldots, M_r$}
                    \State Simulate $(X_{t_j,i})_{j=1,\ldots, m}$
                    \State $S_i \gets  \max_{j=1,\ldots, m} X_{t_j,i} $
                \EndFor
                \State $q_\pm \gets q_{\beta_\pm(M_r,k,\epsilon), M_r}(S_1,\ldots S_{M_r})$
                \State $\widehat{P}^\dagger_\pm \gets 1-\Phi\left( \frac{1 -q_\pm}{\sigma_m} \right)$
                \If{$\widehat{P}^\dagger_+ \leq \alpha$}
                    \State \Return \textsc{safe}
                \ElsIf{$\widehat{P}^\dagger_- \geq \alpha$}
                    \State \Return \textsc{unsafe}
                \EndIf
        \EndFor
        \State \Return \textsc{unsafe}
    \end{algorithmic}
    \label{alg:adaptive-Borell}
\end{algorithm}
\newpage
\subsection{Hybrid scheme}

Algorithm \ref{alg:hybrid} provides Pseudocode for the safety evaluation using the adaptive hybrid scheme described in Section 4.4 of the article.

\begin{algorithm}[H]
\caption{Adaptive hybrid scheme}\label{alg:hybrid}
    \begin{algorithmic}
        \Require Safety threshold: $\alpha>0$, Threshold for confidence intervals $\epsilon>0$, \\
        Discretization $t_1, \dots, t_m$, sequence of sample sizes $0=M_0 < M_1 < \ldots < M_R$,
        Posterior GP: $X_t$
        \For{$r=1,\ldots, R$}
                \State $\beta_\pm \gets \frac{1}{2} \pm \Phi^{-1}(1-\chi(Mk,\epsilon))/\sqrt{4M_r}$ %
                \For{$i=M_{r-1}+1,\ldots, M_r$}
                    \State Simulate $(X_{t_j,i})_{j=1,\ldots, m}$
                    \State $S_i \gets  \max_{j=1,\ldots, m} X_{t_j,i} $
                \EndFor
                \State $\widehat{P} \gets \frac{M_{r-1}}{M_r} \widehat{P} + \frac{1}{M_r} \sum_{i=M_{r-1}+1}^{M_r} \mathbf{1}\left(S_i>1\right)$
                \State $\widehat{P}_{\text{MC}}^+ \gets \widehat{P} + \sqrt{ \frac{2\alpha(1-\alpha)}{M_r} \left| \log \frac{3\epsilon}{\pi^2 r^2} \right| }$
                \State $\widehat{P}_{\text{MC}}^- \gets \widehat{P} - \sqrt{ \frac{2\alpha(1-\alpha)}{M} \left| \log \frac{3\epsilon}{\pi^2 r^2} \right| }$ \Comment{Critical values based on MC scheme}
                \State $q_+ \gets q_{\beta_+(M_r,k,\frac{\epsilon}{2}), M_r}(S_1,\ldots S_{M_r})$
                \State $\widehat{P}^\dagger_+ \gets 1-\Phi\left( \frac{1 -q_\pm}{\tilde{\sigma_m}} \right)$ \Comment{Critical value based on Borell-TIS scheme}
                \If{$\min[\widehat{P}^\dagger_+,\, \widehat{P}_{\text{MC}}^+] \leq \alpha$}
                    \State \Return \textsc{safe}
                \ElsIf{$\widehat{P}_{\text{MC}}^- \geq \alpha$}
                    \State \Return \textsc{unsafe}
                \EndIf
        \EndFor
        \State \Return \textsc{unsafe}
    \end{algorithmic}
    \label{alg:adaptive-Hybrid}
\end{algorithm}

\section{Further information on the examples}

In this section we provide further information on the experiments presented in the main text. We observe the quantities $RMSE$ and $c_h$, as well as the quantities $n_{\text{SAL}}$ which is the number of iterations of the active learning algorithm, and $n_f$, which is defined as the number of training points, that are unsafe due to the ground truth. 
A detailed discussion on this can be found in the respective subsections.

\subsection{Himmelblau's function exploration}

\begin{table}[h]
    \centering
    \caption{Quantities for Himmelblau's function exploration: }
\begin{tabular}{|c|c|c|c|c|}
 \hline 
 method / $\alpha$ & $n_{SAL}$ & RMSE & $c_h$ & $n_{f}$ \\ 
 \hline 
 \hline 
MC / 0.1 & 10.9 $\pm$ 1.7 & 0.2873 $\pm$ 0.0669 & 0.4906 $\pm$ 0.0540 &0.0000 $\pm$ 0.0000 \\ 
 \hline 
AMC (Ours) / 0.1 & 38.2 $\pm$ 7.4 & 0.1553 $\pm$ 0.0543 & 0.7724 $\pm$ 0.0630 &0.2000 $\pm$ 0.6325 \\ 
 \hline 
AB (Ours) / 0.1 & \bf65.1 $\pm$ 7.4 & \bf0.1080 $\pm$ 0.0302 & \bf0.8621 $\pm$ 0.0326 &0.7000 $\pm$ 1.4944 \\ 
 \hline 
ABM (Ours) / 0.1 & 59.6 $\pm$ 9.5 & 0.1511 $\pm$ 0.0801 & 0.8067 $\pm$ 0.0625 &1.1000 $\pm$ 3.1429 \\ 
 \hline 
MC / 0.01 & 10.5 $\pm$ 1.9 & 0.3356 $\pm$ 0.0906 & 0.4476 $\pm$ 0.0711 &0.0000 $\pm$ 0.0000 \\ 
 \hline 
AMC (Ours) / 0.01 & 26.3 $\pm$ 3.9 & 0.2194 $\pm$ 0.0798 & 0.6078 $\pm$ 0.0521 &0.0000 $\pm$ 0.0000 \\ 
 \hline 
AB (Ours) / 0.01 & 50.8 $\pm$ 5.1 & \bf0.1745 $\pm$ 0.0812 & \bf0.7338 $\pm$ 0.0637 &0.0000 $\pm$ 0.0000 \\ 
 \hline 
ABM (Ours) / 0.01 & \bf70.1 $\pm$ 3.0 & 0.1771 $\pm$ 0.0498 & 0.7332 $\pm$ 0.0569 &0.0000 $\pm$ 0.0000 \\ 
 \hline 
MC / 0.001 & 10.6 $\pm$ 2.3 & 0.3482 $\pm$ 0.0648 & 0.4352 $\pm$ 0.0634 &0.0000 $\pm$ 0.0000 \\ 
 \hline 
AMC (Ours) / 0.001 & 31.4 $\pm$ 6.0 & 0.2434 $\pm$ 0.0793 & 0.6457 $\pm$ 0.0556 &0.0000 $\pm$ 0.0000 \\ 
 \hline 
AB (Ours) / 0.001 & 59.1 $\pm$ 10.5 & \bf0.1619 $\pm$ 0.0924 & \bf0.7895 $\pm$ 0.0631 &0.1000 $\pm$ 0.3162 \\ 
 \hline 
ABM (Ours) / 0.001 & \bf64.6 $\pm$ 6.4 & 0.1738 $\pm$ 0.0934 & 0.7470 $\pm$ 0.0887 &0.3000 $\pm$ 0.9487 \\ 
 \hline 
\end{tabular}
    \label{tab:Himmelblau}
\end{table}

\begin{figure}[H]
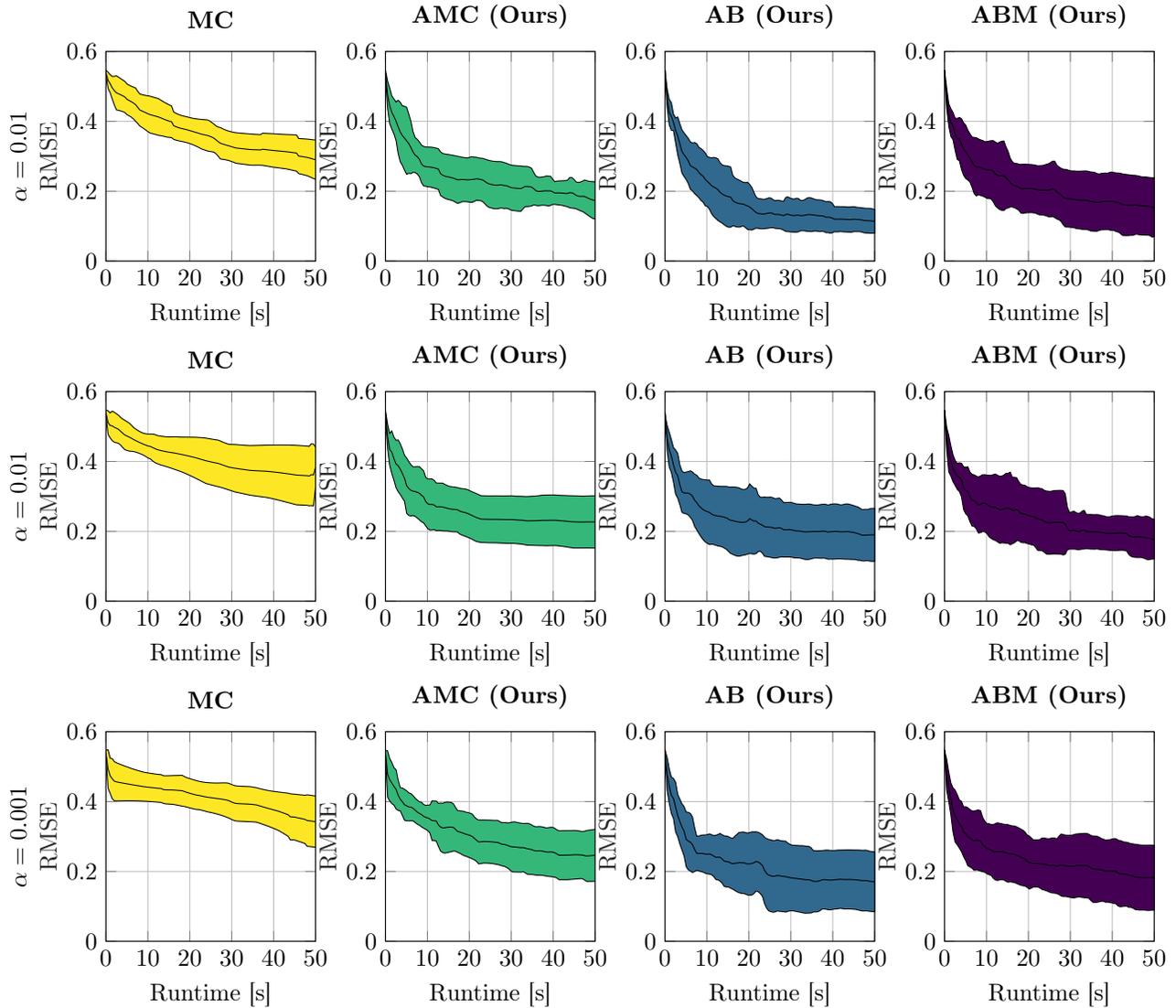

    \begin{center}
    \input{plots/HB_1}
    \input{plots/HB_01.tex}
    \input{plots/HB_001.tex} \\
    \end{center}
    \caption{Himmelblau's function exploration: These plots show the RMSE for three different values of $\alpha$ and the four algorithms. The results are averaged over ten independent seeds and regions are $2\sigma$ confidence intervals.}
    \label{fig:Himmelblau_rmse}
\end{figure}

\newpage

\begin{figure}[H]
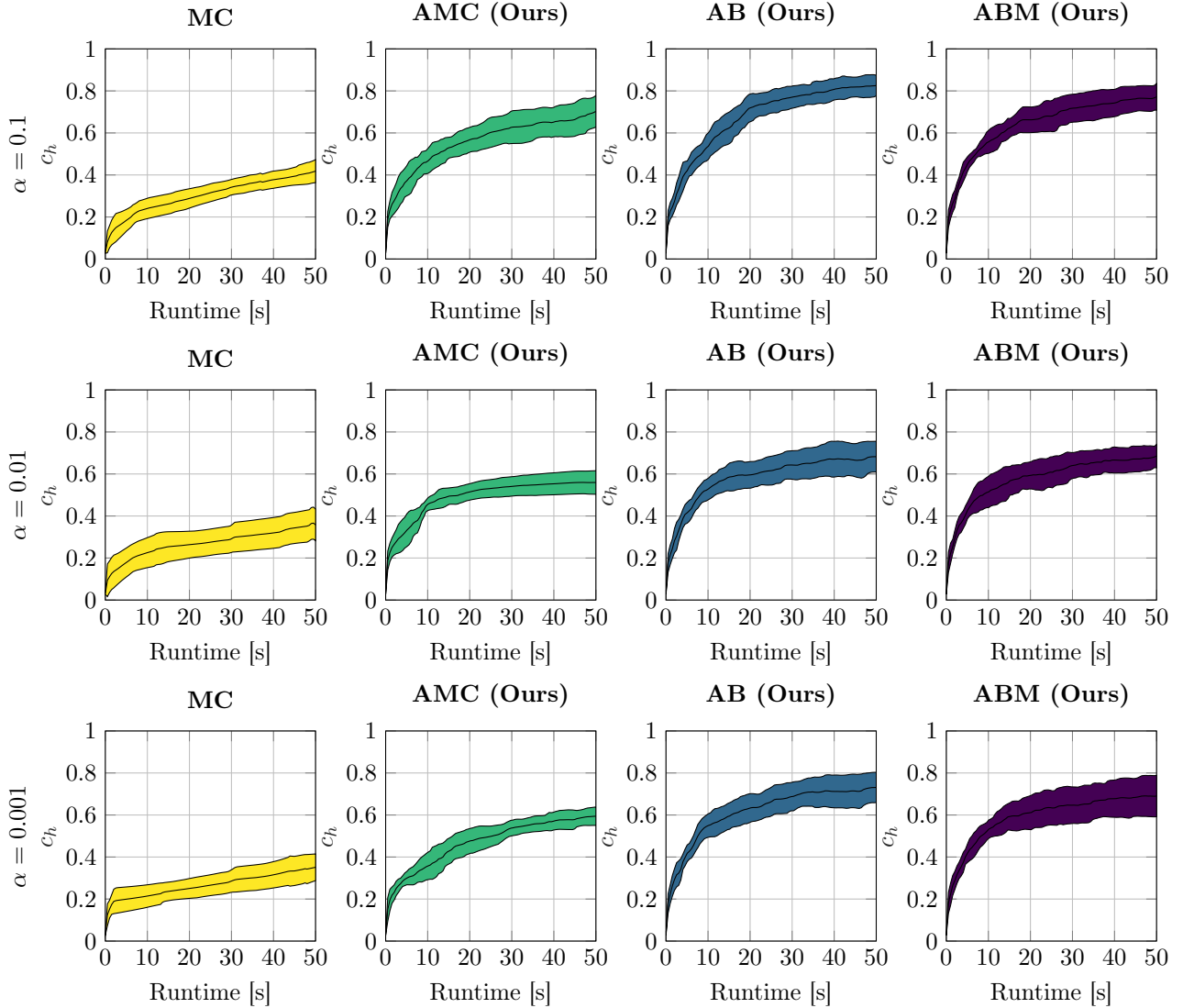

    \begin{center}
    \input{plots/HB_1_ch.tex}
    \input{plots/HB_01_ch.tex} \\
    \input{plots/HB_001_ch.tex} 
    \end{center}
    \caption{Himmelblau function exploration health coverage: These plots show the health coverage $c_h$ for three different values of $\alpha$ and the four algorithms. The results are averaged over ten independent seeds and regions are $2\sigma$ confidence intervals.}
    \label{fig:Himmelblau_ch}
\end{figure}

We use fixed hyperparameters without optimization. The hyperparameters are $\ell_1^2 = \ell_2^2 = 1.0$, $\sigma_f^2=1$ and $\sigma_n = 0.01$. We scale the function with a factor of 0.01 and add normal distributed noise with zero mean and a standard deviation of $0.01$ which coincides with $\sigma_n$. We discretize the trajectories with $m=5$, but only use the endpoint of an explored trajectory as new measurement which is added to the training points.
We present the additional results in Figures \ref{fig:Himmelblau_rmse} and \ref{fig:Himmelblau_ch}, and Table \ref{tab:Himmelblau}.
We see, that our proposed algorithms using the (B.1) bound perform the best with AMC (Ours) also outperforming MC. This results in much more iterations made by the safe active learning algorithm due to fewer samples needed for safety evaluation.

\subsection{Engine control}

\newpage

\begin{table}[H]
    \centering
    \caption{Quantities for engine control}
   \begin{tabular}{|c|c|c|c|c|}
 \hline 
 method / $\alpha$ & $n_{SAL}$ & RMSE & $c_h$ & $n_f$ \\ 
 \hline 
 \hline 
MC / 0.1 & 12.7 $\pm$ 1.4 & 11.6250 $\pm$ 0.2191 & 0.5075 $\pm$ 0.0280 &1.5000 $\pm$ 2.1213 \\ 
 \hline 
AMC (Ours) / 0.1 & 74.8 $\pm$ 9.8 & 10.8945 $\pm$ 0.2759 & 0.6045 $\pm$ 0.0220 &10.1000 $\pm$ 10.3220 \\ 
 \hline 
AB (Ours) / 0.1 & \bf111.0 $\pm$ 7.7 & \bf10.5050 $\pm$ 0.4379 & \bf0.6643 $\pm$ 0.0306 &16.7000 $\pm$ 11.7004 \\ 
 \hline 
ABM (Ours) / 0.1 & 72.7 $\pm$ 9.2 & 10.8287 $\pm$ 0.3261 & 0.6262 $\pm$ 0.0300 &9.1000 $\pm$ 8.5823 \\ 
 \hline 
MC / 0.01 & 12.2 $\pm$ 1.7 & 11.6765 $\pm$ 0.2750 & 0.5007 $\pm$ 0.0358 &1.1000 $\pm$ 1.4491 \\ 
 \hline 
AMC (Ours) / 0.01 & 98.1 $\pm$ 7.2 & 11.2986 $\pm$ 0.1837 & 0.5815 $\pm$ 0.0253 &6.6000 $\pm$ 4.4771 \\ 
 \hline 
AB (Ours) / 0.01 & 129.3 $\pm$ 4.4 & 10.9095 $\pm$ 0.2787 & 0.6258 $\pm$ 0.0215 &5.3000 $\pm$ 2.8304 \\ 
 \hline 
ABM (Ours) / 0.01 & \bf135.9 $\pm$ 3.0 & \bf10.8500 $\pm$ 0.2597 & \bf0.6387 $\pm$ 0.0289 &15.0000 $\pm$ 13.6870 \\ 
 \hline 
MC / 0.001 & 12.5 $\pm$ 1.7 & 11.7215 $\pm$ 0.2253 & 0.4982 $\pm$ 0.0333 &1.2000 $\pm$ 1.5492 \\ 
 \hline 
AMC (Ours) / 0.001 & 94.9 $\pm$ 7.1 & 11.5351 $\pm$ 0.2537 & 0.5600 $\pm$ 0.0356 &6.1000 $\pm$ 6.7897 \\ 
 \hline 
AB (Ours) / 0.001 & 131.1 $\pm$ 2.1 & 11.0656 $\pm$ 0.3132 & 0.6111 $\pm$ 0.0361 &7.8000 $\pm$ 4.0222 \\ 
 \hline 
ABM (Ours) / 0.001 & \bf134.3 $\pm$ 2.8 & \bf10.9203 $\pm$ 0.3193 & \bf0.6161 $\pm$ 0.0240 &10.8000 $\pm$ 6.7626 \\ 
 \hline 
\end{tabular}
    
    \label{tab:Railpressure}
\end{table}
\begin{figure}[H]
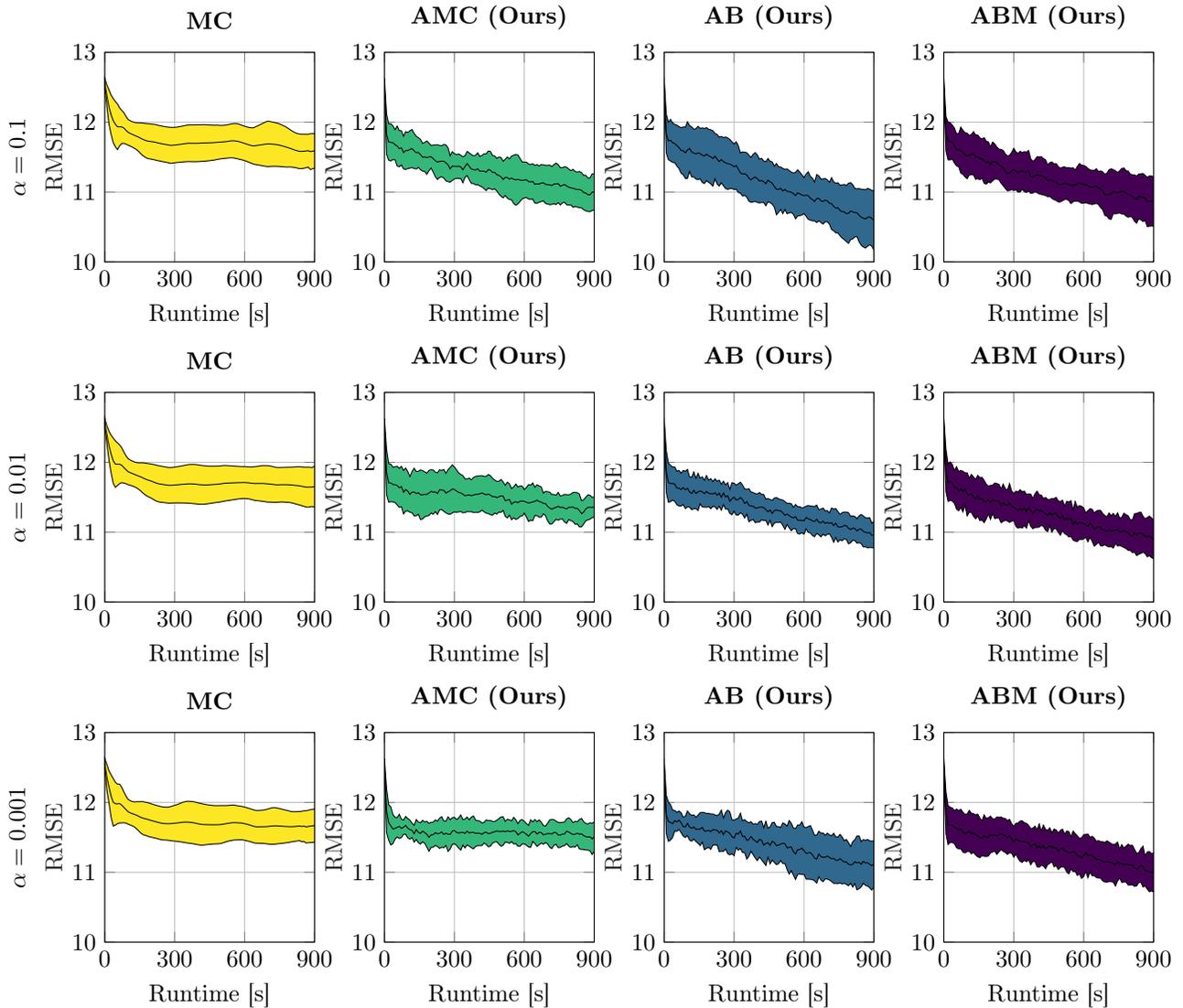

    \begin{center}
    \input{plots/RP_1.tex}
    \input{plots/RP_01.tex} \\
    \input{plots/RP_001.tex} 
    \end{center}
    \caption{Engine control RMSE: These plots show the RMSE for three different values of $\alpha$ and the four algorithms. The results are averaged over ten independent seeds and regions are $2\sigma$ confidence intervals.}
    \label{fig:Railpressure_rmse}
\end{figure}

\begin{figure}[H]
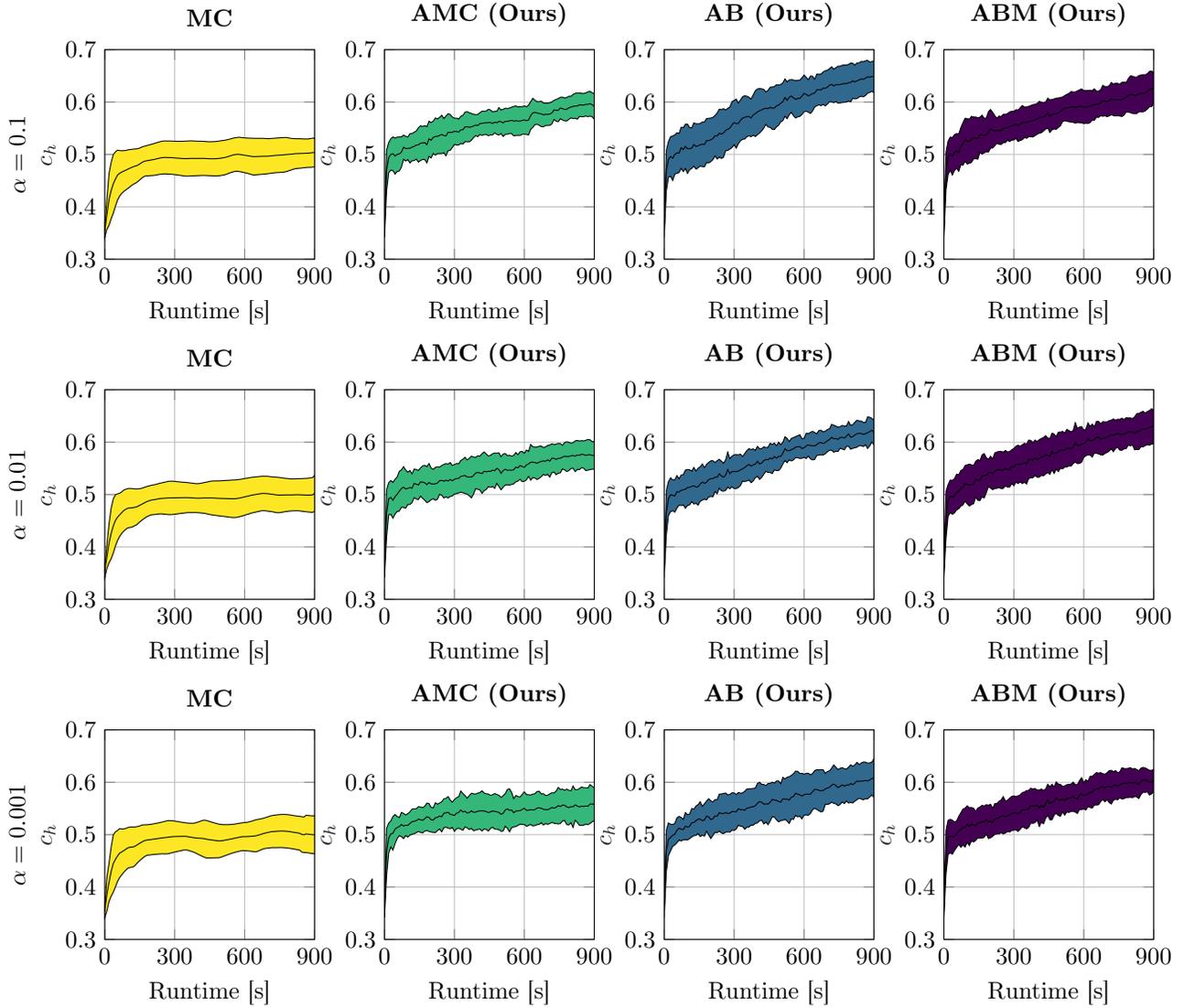

    \begin{center}
    \input{plots/RP_1_ch.tex}
    \input{plots/RP_01_ch.tex} \\
    \input{plots/RP_001_ch.tex} 
    \end{center}
    \caption{Engine control health coverage: These plots show the health coverage $c_h$ for three different values of $\alpha$ and the four algorithms. The results are averaged over ten independent seeds and regions are $2\sigma$ confidence intervals.}
    \label{fig:Railpressure_ch}
\end{figure}

\end{document}